%% file: main.tex
\documentclass[11pt]{article}

\usepackage[]{acl2023}
\usepackage{libertine}      
\usepackage{amsmath}                     
\usepackage{amssymb}
\usepackage[italic]{mathastext}          
\Mathastext
\usepackage{latexsym}
\usepackage[T1]{fontenc}
\usepackage[utf8]{inputenc}
\usepackage[activate={true,nocompatibility},final,tracking=true,kerning=true,spacing=true,factor=1100,stretch=20,shrink=20]{microtype}
\microtypecontext{spacing=nonfrench}
\usepackage{booktabs}
\usepackage{multirow}
\usepackage{graphicx}
\usepackage{xcolor}
\usepackage{url}
\usepackage{hyperref}
\hypersetup{
  colorlinks=true,
  urlcolor=hdrBlue,
  linkcolor=black,
  citecolor=black,
  pdftitle={From Intent to Trajectory: Execution-Grounded Multi-Turn Data Synthesis for OS Agents},
  pdfauthor={Siyuan Luo, Nairong Zheng, Lin Zhou, Tiankuo Yao, Shengyou Yuan, Haojia Yu, Cong Pang, Jiapeng Luo, Lewei Lu}
}
\usepackage{enumitem}
\usepackage{verbatim}
\usepackage{array}
\usepackage{tikz}
\usetikzlibrary{shapes.geometric,arrows.meta,positioning,backgrounds,fit}

\definecolor{hdrBlue}{RGB}{74,126,188}         
\definecolor{hdrTeal}{RGB}{42,158,168}         
\definecolor{hdrPurple}{RGB}{118,98,178}       
\definecolor{hdrOrange}{RGB}{208,112,40}       
\definecolor{hdrGreen}{RGB}{68,148,74}         
\definecolor{hdrGray}{RGB}{96,110,124}         

\definecolor{midBlue}{RGB}{80,138,200}         
\definecolor{midTeal}{RGB}{54,164,176}         
\definecolor{midPurple}{RGB}{128,110,188}      
\definecolor{midOrange}{RGB}{220,118,44}       
\definecolor{midGreen}{RGB}{80,156,86}         
\definecolor{midGray}{RGB}{114,128,142}        

\definecolor{fillBlue}{RGB}{248,251,255}       
\definecolor{fillTeal}{RGB}{247,252,253}       
\definecolor{fillPurple}{RGB}{250,249,255}     
\definecolor{fillOrange}{RGB}{255,250,245}     
\definecolor{fillGreen}{RGB}{247,253,247}      
\definecolor{fillGray}{RGB}{250,251,252}       

\definecolor{deltaUp}{RGB}{40,130,60}          
\definecolor{deltaDn}{RGB}{198,58,42}          
\newcommand{\up}[1]{\rlap{\textsuperscript{\textcolor{deltaUp}{\tiny\,$+$#1\%}}}}   
\newcommand{\dn}[1]{\rlap{\textsuperscript{\textcolor{deltaDn}{\tiny\,$-$#1\%}}}}   

\usepackage{algorithm}
\usepackage{algpseudocode}
\usepackage{float}
\usepackage{tcolorbox}
\tcbuselibrary{skins}
\usepackage{placeins}  
\usepackage{balance}   
\usepackage[skip=8pt,belowskip=3pt,font=small]{caption}  

\usepackage[scaled=0.96,varqu,varl]{zi4}
\makeatletter
\renewcommand{\@dblfptop}{0pt plus 0fil}
\renewcommand{\@dblfpsep}{6pt plus 1fil}
\renewcommand{\@dblfpbot}{0pt plus 0fil}
\makeatother

\newcommand{\RESULT}[1]{}
\newcommand{\DELTA}[1]{}
\newcommand{\N}[1]{}

\title{ISE: An Execution-Grounded Recipe for\\Multi-Turn OS-Agent Trajectories}

\author{
  Siyuan Luo\textsuperscript{1,*} \quad Nairong Zheng\textsuperscript{2,*} \quad Lin Zhou\textsuperscript{2,\dag} \quad Tiankuo Yao\textsuperscript{2,\dag} \quad Shengyou Yuan\textsuperscript{2,\dag} \\
  \textbf{Haojia Yu}\textsuperscript{2} \quad \textbf{Cong Pang}\textsuperscript{2} \quad \textbf{Jiapeng Luo}\textsuperscript{2} \quad \textbf{Lewei Lu}\textsuperscript{2,\ddag} \\
  \vspace{0.3em}
  {\normalfont\normalsize \textsuperscript{1}University of Electronic Science and Technology of China \quad \textsuperscript{2}SenseTime Research} \\
  {\normalfont\normalsize \texttt{valierelane@gmail.com} \quad \texttt{luotto@sensetime.com}}
}

\begin{document}
\flushbottom  

\renewcommand{\topfraction}{0.85}
\renewcommand{\bottomfraction}{0.6}
\renewcommand{\textfraction}{0.12}
\renewcommand{\floatpagefraction}{0.75}
\renewcommand{\dbltopfraction}{0.85}
\renewcommand{\dblfloatpagefraction}{0.75}
\setcounter{topnumber}{2}
\setcounter{bottomnumber}{2}
\setcounter{totalnumber}{3}
\setcounter{dbltopnumber}{2}
\makeatletter
\@afterindentfalse
\def\paragraph{\@startsection{paragraph}{4}{\z@}%
   {1.5ex plus 0.5ex minus .2ex}{-1em}{\normalsize\bfseries}}
\def\subparagraph{\@startsection{subparagraph}{5}{\z@}%
   {1.5ex plus 0.5ex minus .2ex}{-1em}{\normalsize\bfseries}}
\makeatother
\maketitle

\makeatletter
\def\blfootnote#1{%
  \begingroup
    \long\def\@makefntext##1{\noindent##1}%
    \renewcommand\thefootnote{}\footnote{#1}%
    \addtocounter{footnote}{-1}%
  \endgroup
}
\makeatother
\blfootnote{\makebox[0.9em][l]{\textsuperscript{*}}Co-first authors. \\ \makebox[0.9em][l]{\textsuperscript{\dag}}Core contributors. \\ \makebox[0.9em][l]{\textsuperscript{\ddag}}Corresponding author.}

\begin{abstract}
\setlength{\parindent}{1em}
\indent\input{sections/abstract}
\end{abstract}

\input{sections/introduction}
\input{sections/related_work}
\input{sections/dataset_analysis}
\input{sections/method}
\input{sections/experiments}
\input{sections/limitations}
\input{sections/conclusion}

\balance  
\bibliography{refs}

\end{document}

%% file: sections/abstract.tex
Training capable OS agents requires data that simultaneously captures structured user intents, multi-turn task delegation, and grounded tool execution---properties absent from existing datasets. We propose \textsc{ISE} (\textbf{I}ntent $\rightarrow$ \textbf{S}imulate $\rightarrow$ \textbf{E}xecute), a three-stage synthesis paradigm that addresses these gaps jointly.

Stage~1 constructs $\sim$50{,}000 structured intents via a 4D framework (Persona $\times$ Domain $\times$ Task $\times$ Complexity); after deduplication the pool contains $43{,}956$ unique intents and attains a Vendi Score of $61.57$ over the \emph{entire} pool on \textit{mpnet-base-v2} embeddings (cosine kernel, $q{=}1$). Stage~2 drives multi-turn user--agent interaction through a role-locked user simulator that grounds each user turn in actual execution outcomes, producing 23{,}132 complete trajectories averaging 8.12 user turns and 68.24 total dialogue turns. Stage~3 executes every tool call in a live, isolated OS workspace, yielding authentic failure--recovery dynamics rather than simulated responses.

Fine-tuning on \textsc{ISETrace} lifts ClawEval pass@1 from 19.3 to 37.7 on Qwen3-8B (agent tool-use tasks, common-denominator protocol), surpassing both a GPT-4o zero-shot reference and a $4\times$-larger Qwen3-32B base; a Stage~2 ablation indicates multi-turn simulation contributes a substantial share of the gain. We release all code and data at \url{https://github.com/Valiere01/ISE-Trace}.

%% file: sections/introduction.tex
\section{Introduction}
\label{sec:intro}

Large language model agents are increasingly deployed in stateful operating-system environments, yet the training data used to teach them still underrepresents four properties of real use: user intents are implicit and underspecified, actions have external side effects, users react to partial progress and failure, and successful completion is often verifiable only through environment state. Despite rapid progress in large language models, agents still fail on more than half of realistic multi-turn OS tasks~\citep{yao2024taubench}, and the bottleneck is not model capacity---it is training data.

A closer look at current synthesis pipelines reveals three systematic structural gaps. \textbf{Gap 1 (Intent-first bias):} Most pipelines start from a list of available APIs or tools---e.g., the 16k+ REST endpoints on RapidAPI or a curated SDK catalog~\citep{qin2024toolbench,liu2024apigen}---and \emph{back-derive} tasks from each tool (``\emph{get\_weather(city)}'' $\rightarrow$ ``What's the weather in Tokyo?''). The resulting task distribution therefore mirrors the catalog rather than what users actually want; long-tail and cross-tool intents are systematically under-represented. The natural alternative---asking an LLM to free-generate user tasks---fares no better: instruction-tuned LLMs exhibit a well-documented \emph{mode collapse} toward the high-frequency phrasings they have seen most often~\citep{wang2023selfinstruct} (algorithmic puzzles, generic email templates, customer-service openers), producing tasks that look diverse on the surface but cluster in a narrow region of intent space. \textbf{Gap 2 (Single-turn bias):} Nearly all OS agent datasets are single-turn~\citep{sun2024osgenesis,xu2025agenttrek}, failing to capture the multi-turn task delegation, correction, and verification cycles central to real agent interactions. Even pipelines with user simulators~\citep{prabhakar2025apigenmt,chen2026cove} suffer from \emph{role drift}---instruction-tuned LLMs gradually adopt assistant-style language---and \emph{state hallucination}---simulators issue follow-up requests based on assumed states that diverge from actual OS state~\citep{zhou2026sim2real}. \textbf{Gap 3 (Simulated execution):} Tool execution is typically simulated rather than real~\citep{mitra2024agentinstruct,chen2026eigendata}, training agents on a hallucinated execution distribution that diverges from actual OS behavior and producing almost no authentic failure-recovery examples.

These gaps compound: missing any one of them produces training data that is unrepresentative, limited, or disconnected from real execution semantics.

\begin{figure*}[!tbp]
\centering
\includegraphics[width=0.95\linewidth]{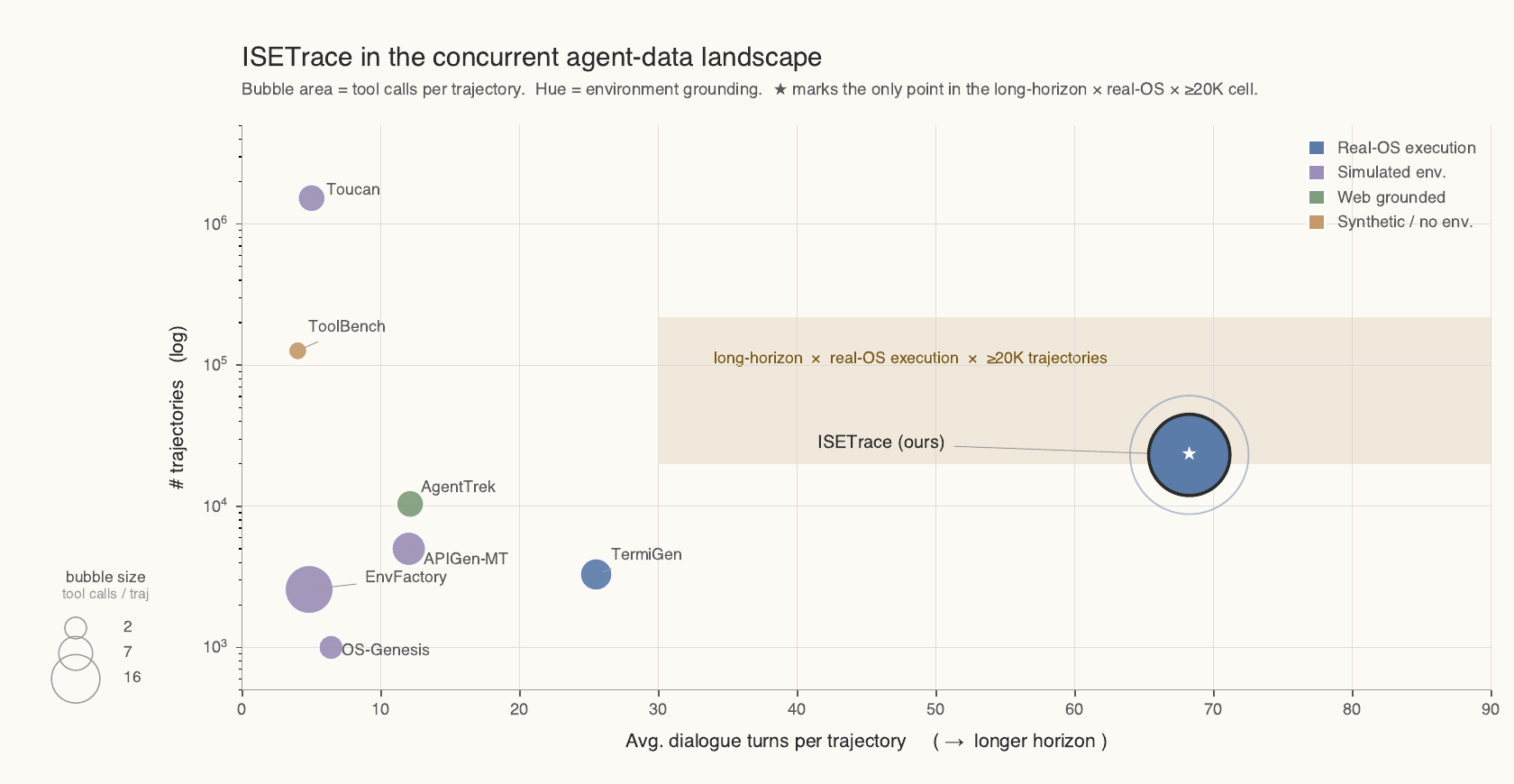}
\caption{\textsc{ISETrace} in the concurrent agent-data landscape. Each circle is one corpus (axis = avg.\ dialogue turns per trajectory; y-axis = \#trajectories on log scale). Bubble area encodes tool calls per trajectory; hue encodes environment grounding (real-OS / simulated / web / synthetic). The shaded band marks the long-horizon $\times$ real-OS execution $\times$ $\geq$20K trajectories regime, which \textsc{ISETrace} alone occupies among concurrent works.}
\label{fig:landscape}
\end{figure*}

We propose \textbf{ISE} (\textbf{I}ntent $\rightarrow$ \textbf{S}imulate $\rightarrow$ \textbf{E}xecute), a three-stage synthesis paradigm that addresses all three gaps jointly. Figure~\ref{fig:landscape} situates \textsc{ISETrace} against concurrent work.
\textbf{Stage~1} constructs $\sim$50{,}000 structured intents by independently sampling four axes---Persona, Domain subset, Task sequence, Complexity---and then expanding the chosen tasks into their required tool set: on average each intent spans 2.35 domains and 4.40 ordered tasks, which together invoke 3.18 distinct tools (a derived statistic, not a fifth sampling axis). After deduplication the pool contains $43{,}956$ unique intents and attains a Vendi Score of $61.57$ on \textit{mpnet-base-v2} embeddings (cosine, $q{=}1$) computed over the \emph{entire} pool.
\textbf{Stage~2} drives multi-turn interaction through a role-locked user simulator with four behavioral constraints that suppress role drift and state hallucination, producing 23{,}132 complete trajectories with 91.1\% containing 6--10 user turns (avg.\ 8.12 user turns, 68.24 total dialogue turns).
\textbf{Stage~3} grounds all tool calls in real OS execution in isolated live workspaces, ensuring trajectories reflect authentic OS behavior rather than simulated tool responses.

\paragraph{Contributions.}
\begin{enumerate}[leftmargin=*,labelindent=0pt,labelsep=0.4em,itemsep=2pt]
\item \textbf{ISE paradigm and \textsc{ISETrace} dataset}: a three-stage recipe and the resulting multi-turn OS-agent trajectory corpus, synthesized from structured intents and grounded in live OS execution.
\item \textbf{Diversity and ablation evidence}: full-stack diversity quantification (embedding, lexical, structural) and an ablation isolating the contribution of multi-turn simulation (\S\ref{sec:exp:diversity},~Table~\ref{tab:ise_ablation}).
\end{enumerate}

%% file: sections/related_work.tex
\section{Related Work}
\label{sec:related}

\subsection{Agentic Data Synthesis}

\paragraph{Tool-first vs.\ environment-driven synthesis.}
Tool-first pipelines~\citep{qin2024toolbench,liu2024apigen,mitra2024agentinstruct} back-derive tasks from API catalogs, mirroring tool space rather than user-need space and operating without live execution. Environment-driven pipelines instead infer tasks post-hoc from random GUI exploration~\citep{sun2024osgenesis} or web tutorials~\citep{xu2025agenttrek}, with coverage bounded by the seed pool and no multi-turn user simulation. ISE inverts both: structured 4D intent sampling prospectively drives synthesis from user-need space (diversity quantified in \S\ref{sec:exp:diversity}).

\paragraph{Multi-turn synthesis and verification.}
The closest competitor, \citet{chen2026eigendata}, synthesizes multi-turn tool-use data with per-instance LLM-written checkers; ISE instead uses real OS execution (a physically deterministic signal) plus role-locked simulation that grounds every user turn in execution state. \citet{chen2026cove} and \citet{prabhakar2025apigenmt} add user simulation or LLM-committee verification but over simulated API environments, while \citet{zhu2026termigen}, \citet{lin2026cligym}, and \citet{yang2025toolmind} pursue execution-based environments or evaluation without multi-turn user simulation.

\paragraph{Concurrent 2026 work.}
Several concurrent efforts target tool-use or MCP environments. Toucan~\citep{wang2025toucan} synthesizes 1.5M trajectories from $\sim$500 MCP servers, of which 567,262 (37\%) are multi-turn. EnvFactory~\citep{minrui2026envfactory} generates 2{,}575 trajectories from 85 verified environments with an average of 4.82 turns and 3.29 steps per turn. COVERT~\citep{anonymous2026covert} focuses on oracle-preserving RL augmentations and reports BFCL v3 / ACEBench accuracy rather than corpus-level statistics. A parallel line of GUI-centric work (OpenMobile~\citep{anonymous2026openmobile}, ToolCUA~\citep{anonymous2026toolcua}, CUA-Gym~\citep{anonymous2026cuagym}, Video2GUI~\citep{anonymous2026wildgui}) targets visual interaction rather than shell semantics. Our work differs along two axes that the corpora above do not jointly cover: (i)~all trajectories execute against a real shell, and (ii)~we report embedding-level diversity (Vendi / Self-BLEU / Distinct-N) alongside the corpus. Table~\ref{tab:related_positioning} summarizes the comparison; whether longer per-trajectory length translates into downstream gains is left to \S\ref{sec:exp} rather than asserted here.

\subsection{Agent Training Paradigms}

SFT on synthetic trajectories~\citep{zeng2024agenttuning,shi2025taskcraft} remains the dominant paradigm for OS agent training and is the regime we evaluate. We deliberately separate \emph{data composition} (the contribution of this work) from training-algorithm choices: holding the base model and training objective fixed, the question is whether 4D structured intents, role-locked multi-turn simulation, and execution grounding move the needle.

\subsection{Multi-Turn Evaluation}

\citet{yao2024taubench} provide the standard multi-turn benchmark with an LLM user simulator. \citet{zhou2026sim2real} show LLM simulators are systematically more cooperative and stylistically uniform than real users---directly motivating our role-locking design. \citet{liu2024agentbench} provide broader OS-level evaluation. Table~\ref{tab:related_positioning} positions ISETrace against twelve contemporary corpora spanning 2023--2026.

\begin{table*}[!tbp]
\centering
\small
\caption{Positioning of ISETrace (ours) against twelve contemporary agent-trajectory corpora. \textbf{Turns}: average total turns per trajectory; \textbf{Tools/T}: average tool calls per trajectory; \textbf{Toks}: average tokens per trajectory (k=thousand); \textbf{MT}: multi-turn user simulation---presence of a dynamic user simulator whose follow-up turns are conditioned on intermediate execution outcomes (not multi-step agent execution alone); \textbf{Real}: real OS execution (vs.\ simulated/GUI sandbox). \checkmark~= yes; $\sim$~= partial; \texttimes{}~= no; ``--''~= original paper does not report. In the \textbf{Real} column, a parenthesized tag, where present, names the execution substrate (\textit{sim}=simulated, \textit{MCP}=MCP servers, \textit{mobile}/\textit{GUI}=device/GUI sandbox, \textit{mock}=mocked I/O, \textit{offline}=replayed logs, \textit{full OS}=live operating system). All numbers verified against source PDFs. $^{\ddagger}$~Measured by us from the released dataset under our own tool-call protocol (the source paper does not report this figure); see \S\ref{sec:exp:diversity}.}
\label{tab:related_positioning}
\resizebox{\textwidth}{!}{%
\begin{tabular}{llrrrrcc}
\toprule
\textbf{Work} & \textbf{Year} & \textbf{\#Traj} & \textbf{Turns} & \textbf{Tools/T} & \textbf{Toks} & \textbf{MT} & \textbf{Real} \\
\midrule
ToolBench~\citep{qin2024toolbench} & 2023 & 126{,}486 & 4.0 & -- & -- & \texttimes & $\sim$ \\
OS-Genesis~\citep{sun2024osgenesis} & 2024 & 1{,}000 & 6.4 & -- & -- & \texttimes & $\sim$ \\
AgentTrek~\citep{xu2025agenttrek} & 2024 & 10{,}398 & 12.1 & -- & -- & \texttimes & $\sim$ \\
APIGen-MT~\citep{prabhakar2025apigenmt} & 2025 & 5{,}000 & 12 & 4.39$^{\ddagger}$ & -- & \checkmark & $\sim$ (sim) \\
TermiGen~\citep{zhu2026termigen} & 2026-01 & 3{,}291 & 25.5 & -- & 8.7k & \texttimes & \checkmark \\
Toucan~\citep{wang2025toucan} & 2025-10 & 1{,}527{,}259 & 15.8$^{\ddagger}$ & 3.42$^{\ddagger}$ & -- & $\sim$ (37\%) & $\sim$ (MCP) \\
EnvFactory~\citep{minrui2026envfactory} & 2026-05 & 2{,}575 & 4.82 & -- & -- & $\sim$ & $\sim$ (MCP) \\
COVERT~\citep{anonymous2026covert} & 2026-04 & -- & -- & -- & -- & \texttimes & \checkmark \\
OpenMobile~\citep{anonymous2026openmobile} & 2026-04 & 2{,}800 & -- & -- & -- & \texttimes & $\sim$ (mobile) \\
ToolCUA~\citep{anonymous2026toolcua} & 2026-05 & 10{,}000 & -- & -- & -- & \texttimes & $\sim$ (GUI) \\
CUA-Gym~\citep{anonymous2026cuagym} & 2026-05 & 32{,}112 & -- & -- & -- & \texttimes & \checkmark (mock) \\
Video2GUI~\citep{anonymous2026wildgui} & 2026-05 & 12.7M & 9.7 & -- & -- & \texttimes & \texttimes (offline) \\
\midrule
\textbf{ISETrace (Ours)} & 2026 & 23{,}132 & \textbf{68.24} & \textbf{29.26} & \textbf{50.5k} & \checkmark & \checkmark (\textbf{full OS}) \\
\bottomrule
\end{tabular}%
}
\end{table*}

%% file: sections/dataset_analysis.tex
\section{\textsc{ISETrace} Dataset Analysis}
\label{sec:data_analysis}
\label{sec:exp:diversity}

We characterize the dataset along three orthogonal axes---semantic (embedding), lexical (n-gram), and structural (tool-call topology)---to verify that 4D sampling combined with execution grounding produces qualitatively richer trajectories than tool-first or single-turn alternatives.

\paragraph{Embedding diversity: Vendi Score.}
We compute the Vendi Score~\citep{friedman2022vendi} (order $q{=}1$, cosine kernel) over \textit{all-mpnet-base-v2} embeddings\footnote{Hugging Face model id: \textit{sentence-transformers/\allowbreak{}all-mpnet-base-v2}.}. The intent pool contains $43{,}956$ unique intents after deduplication; we evaluate Vendi both at the conventional $N{=}500$ subsample (for direct comparability with prior work) and over the \emph{entire} pool. Computation at full $N$ is made tractable by the identity $\mathrm{spec}(XX^{\top}) = \mathrm{spec}(X^{\top}X)$ on the non-zero eigenvalues, which reduces the kernel eigendecomposition from an $N\times N$ to a $768\times 768$ matrix. \textsc{ISETrace} attains a Vendi Score of $51.27 \pm 1.49$ at $N{=}500$ (30 bootstraps) and $\mathbf{61.57}$ over the full pool. Table~\ref{tab:vendi} reports the per-configuration breakdown at $N{=}500$, showing the score is robust across domain and persona slices and only drops noticeably under a single-industry restriction (Tech-only, $41.61$).

\begin{table}[!t]
\centering
\small
\caption{Vendi Score breakdown. Top row reports the full-pool figure ($N=43{,}956$); subsequent rows are at $N{=}500$ for direct comparability with prior work. The score is stable across multi- vs.\ single-domain and cross- vs.\ single-industry persona slices, but contracts when restricted to a single industry (Tech-only).}
\label{tab:vendi}
\resizebox{\columnwidth}{!}{%
\begin{tabular}{lcc}
\toprule
\textbf{Configuration} & \textbf{$N$} & \textbf{Vendi Score} \\
\midrule
\textsc{ISETrace} (full pool)                  & 43{,}956 & \textbf{61.57} \\
\midrule
\textsc{ISETrace} (random subsample)           & 500 & 51.27 $\pm$ 1.49 \\
\quad Multi-domain only                        & 500 & 51.50 \\
\quad Single-domain only                       & 500 & 51.40 \\
\quad Cross-industry persona                   & 500 & 51.60 \\
\quad Single-industry (Tech only)              & 500 & 41.61 \\
\bottomrule
\end{tabular}%
}
\end{table}

\paragraph{Lexical diversity.}
On a length-normalised distinct-n protocol (lowercased, whitespace-tokenised, truncated to the first $K$ tokens, $N{=}5{,}000$ samples per corpus),\footnote{For \textsc{ISETrace} the scored text is the structured \emph{natural\_language\_intent} (mean $124$ words); for the instruction baselines it is the native instruction / first user turn (e.g.\ CodeAlpaca and Alpaca average ${\sim}15$ words). The first-$K$-token truncation ($K{\in}\{20,50\}$) normalises for this length gap; at $K{=}50$ the short-instruction corpora (CodeAlpaca, Alpaca) have too few $\geq\!50$-token examples and are omitted, leaving \textsc{ISETrace} alongside the longer-form ShareGPT and WizardLM.} ISETrace's lexical diversity is comparable to public instruction corpora such as ShareGPT (Vicuna split)~\citep{chiang2023vicuna} and WizardLM Evol-Instruct~\citep{xu2024wizardlm}, confirming that its intents are non-templated rather than rephrasings of a small seed set. We therefore do not claim dominant lexical diversity; the decisive cross-corpus gains appear on the embedding axis (Vendi, above) and on tool-call structure (below). The length gap itself is substantive rather than incidental: each \textsc{ISETrace} intent encodes a multi-step composite workload---on average $4.40$ tasks spanning $2.35$ domains, with $95.5\%$ of intents carrying concrete numeric parameters (thresholds, quantities, identifiers)---whereas single-sentence instruction corpora pose one atomic request per example. Lexical length here is a symptom of task complexity, not padding.

\begin{figure*}[!tbp]
\centering
\includegraphics[width=\linewidth]{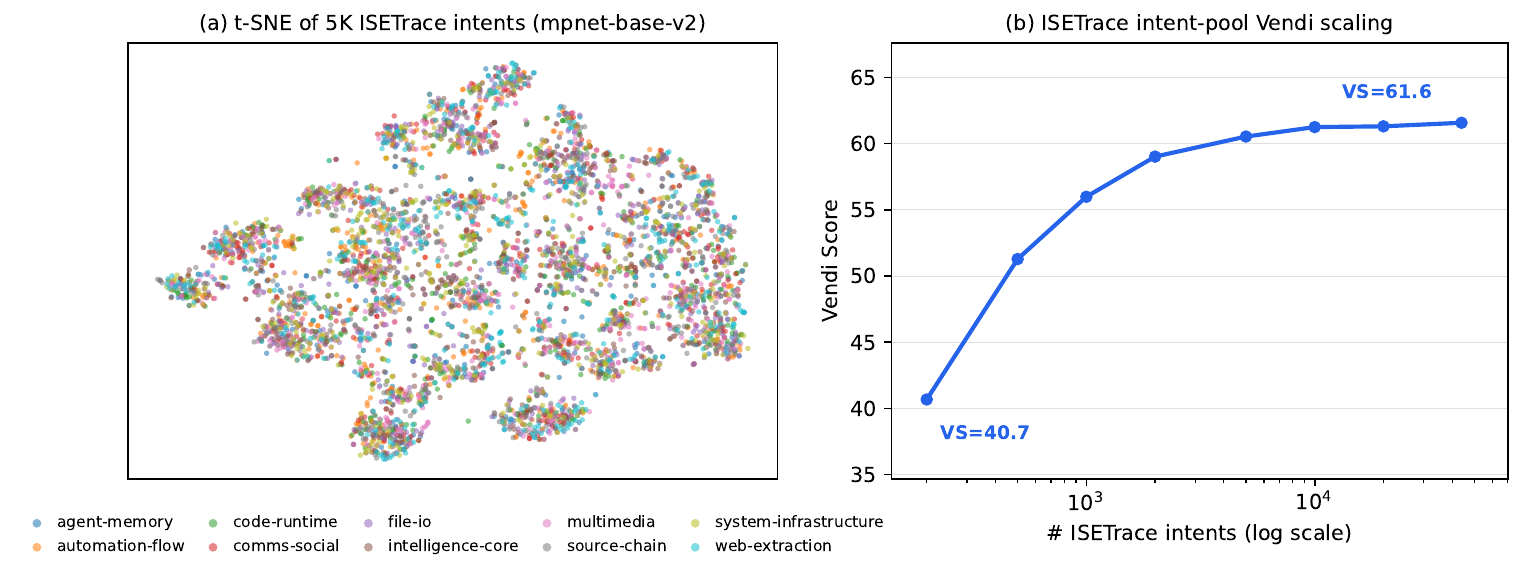}
\caption{\textsc{ISETrace} coverage analysis. \textbf{Left}: t-SNE projection of $5{,}000$ sampled intents (mpnet-base-v2 embeddings), colored by primary domain---spread is broad with all 10 domains overlapping rather than clustered. \textbf{Right}: Vendi scaling curve over $N\in\{200,\dots,43{,}956\}$ (log $N$), growing monotonically from $40.67$ to $61.57$ at the full pool; the marginal gain decays to $+0.27$ in the last interval, indicating the pool is close to but has not reached saturation.}
\label{fig:coverage}
\end{figure*}

\paragraph{Coverage projection.}
Figure~\ref{fig:coverage} (left) shows a t-SNE projection of 5{,}000 \textsc{ISETrace} intents colored by primary domain; the embedding occupies a broad spread with all 10 domains overlapping rather than forming isolated clusters. The right panel plots the Vendi Score against sample size $N\in\{200, 500, 1{,}000, 2{,}000, 5{,}000, 10{,}000, 20{,}000, 43{,}956\}$ (the last point being the full deduplicated pool): the curve increases monotonically from $40.67$ ($N{=}200$) to $61.57$ (full pool), with the marginal gain falling from $+10.6$ between $N{=}200$ and $N{=}500$ to $+0.27$ between $N{=}20{,}000$ and the full pool. The pool is therefore close to but has not reached saturation, evidencing that the synthesis pipeline keeps producing genuinely new content rather than rephrasings of a fixed seed pool.

\begin{figure}[!bp]
\centering
\includegraphics[width=\columnwidth]{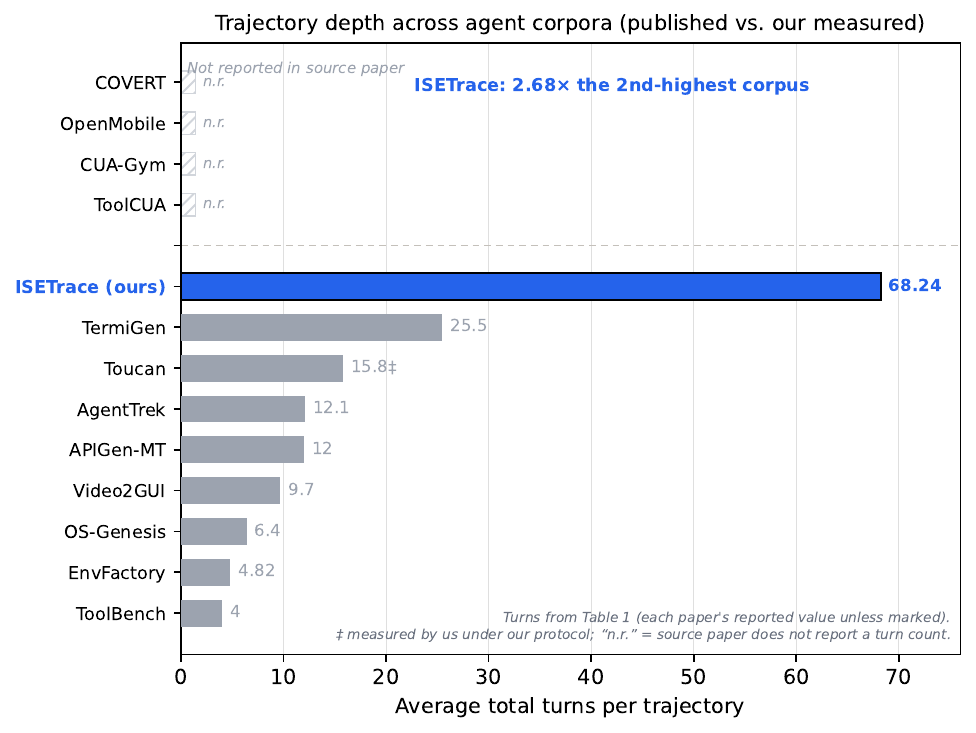}
\caption{Trajectory depth across thirteen agent corpora: average total turns per trajectory. \textsc{ISETrace} (ours, highlighted) is the deepest by a wide margin at $68.24$ turns, $2.68\times$ the next-highest corpus. Values are from Table~\ref{tab:related_positioning}; hatched ``n.r.'' bars denote corpora whose source paper does not report a turn count (shown rather than dropped to avoid selection bias).}
\label{fig:cross_corpus_turns}
\end{figure}

\paragraph{Cross-dataset comparison.}
We place \textsc{ISETrace} on the broader landscape of public agent SFT corpora along the axis that most directly reflects interaction richness: average trajectory depth (total turns per trajectory; Figure~\ref{fig:cross_corpus_turns}). However, the Table~\ref{tab:related_positioning} turn counts are each corpus's self-reported figure under heterogeneous definitions (reasoning traces, GUI steps, dialogue turns) and are not strictly commensurable, so we defer the decisive comparison to a single common protocol below.

\paragraph{Same-protocol re-measurement.}
To remove the definitional ambiguity for the two baselines that release complete multi-turn \emph{conversational} trajectories, we re-measure them under the exact protocol used for our own---total messages per trajectory, counting every user, assistant, and tool-result message. On the full APIGen-MT-5k release~\citep{prabhakar2025apigenmt} ($N{=}5{,}000$) we measure $18.5$ messages/trajectory (the paper self-reports ``12 turns''), and on the Toucan multi-turn subset~\citep{wang2025toucan} ($N{=}35{,}243$) we measure $15.8$. Under this single common protocol \textsc{ISETrace} ($68.24$) is $3.7\times$ and $4.3\times$ deeper, respectively---the direct corpus-level signature of the role-locked user simulator (\S\ref{sec:method:rlus}), which sustains long exchanges by grounding each user turn in execution outcomes rather than terminating after one request--response pair. We restrict this re-measurement to these two corpora because the remaining public datasets either release single-step SFT samples that do not reconstruct trajectory-level turns (e.g.\ AgentTrek, OpenMobile), record non-conversational GUI action sequences (e.g.\ Video2GUI, OS-Genesis), or release no trajectory data; for those, the self-reported value in Table~\ref{tab:related_positioning} is retained.

As a complementary check we apply the identical Vendi protocol (mpnet-base-v2, cosine, $q{=}1$) to the first user-role message of \textsc{ISETrace} (23K), APIGen-MT-5k~\citep{prabhakar2025apigenmt}, AgentTrek~\citep{xu2025agenttrek}, and Toucan-1.5M~\citep{xu2025toucan} (4{,}000-trajectory cap). \textsc{ISETrace} reaches Vendi $97$ at $N{=}4{,}000$ ($3.5\times$ APIGen-MT-5k's $28$, on par with AgentTrek's $110$, below Toucan's $147$); since this surface measures simulator-rewritten first-message text rather than the underlying intent, we do not read it as a diversity claim---the decisive cross-corpus gap remains the trajectory-depth axis above.

\begin{figure}[!tbp]
\centering
\includegraphics[width=\columnwidth]{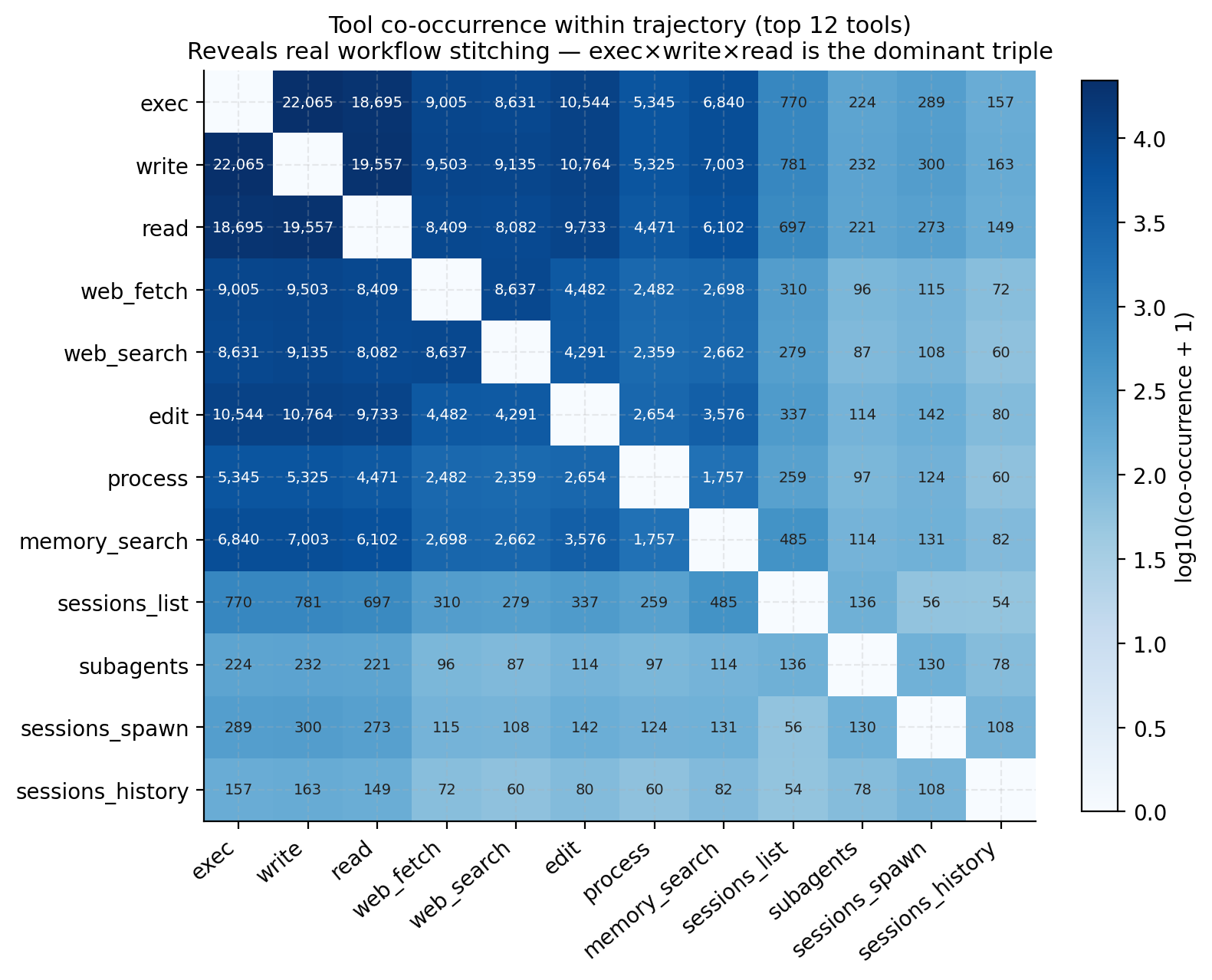}
\caption{Pairwise tool co-occurrence within trajectories (top 12 of 16 tools, $\log_{10}$ scale; aggregated over all $676{,}901$ tool calls in the $23{,}132$ released trajectories). The \textit{exec}--\textit{write}--\textit{read} triangle dominates (\textit{exec}$\times$\textit{write}$=22.1$K, \textit{write}$\times$\textit{read}$=19.6$K, \textit{exec}$\times$\textit{read}$=18.7$K), reflecting the iterative ``write-script $\to$ run $\to$ inspect'' workflow the trigrams above make explicit.}
\label{fig:tool_cooccurrence}
\end{figure}

\paragraph{Structural diversity: tool-call topology.}
Trajectories average 29.26 tool calls drawn from 4.69 unique tools out of 16. The top three trigrams of consecutive tool calls---\textit{exec--exec--exec} (126.8K occurrences); \textit{write--exec--exec} (33.6K); \textit{exec--write--exec} (29.0K); together with \textit{web\_fetch--web\_fetch--web\_fetch} (22.0K)---reflect real engineering patterns (iterative scripting, write-and-test, crawl chains), not generic single-step query/response. Figure~\ref{fig:tool_cooccurrence} visualizes the tool co-occurrence matrix.

%% file: sections/method.tex
\section{ISE: Synthesis Paradigm}
\label{sec:method}

\subsection{Overview}
\label{sec:method:overview}

We instantiate ISE on top of OpenClaw, a production agent platform providing a unified tool API, live OS execution, and reproducible workspace isolation. The paradigm is agent-system-agnostic: any platform supporting live tool execution and workspace isolation can serve as the execution substrate. Dataset statistics are summarized in Table~\ref{tab:dataset_stats}.

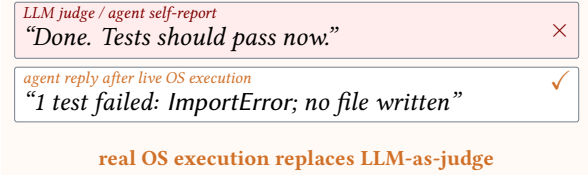
\begin{figure*}[!tbp]
\centering
\resizebox{\linewidth}{!}{\input{figures/fig_pipeline_v3.tex}}
\caption{The ISE synthesis paradigm at a glance. Each of the three stages contrasts a typical failure mode of prior work (top, $\times$) with what ISE contributes (bottom, $\checkmark$). \emph{(I)~Intent}: structured 4D sampling over $\mathcal{P}\times 2^{\mathcal{D}}\times\mathcal{T}^{*}\times\mathcal{C}$ produces a diverse intent pool. \emph{(S)~Simulate}: a role-locked user simulator enforces four behavioral principles (perspective lock, register matching, incremental advancement, responsive conditioning) that suppress role drift. \emph{(E)~Execute}: the agent runs every turn against a real OS, and its post-execution reply is fed back to the simulator, replacing the LLM-as-judge / agent self-report loop. A walked-through trajectory instance is shown in Figure~\ref{fig:example}.}
\label{fig:pipeline}
\end{figure*}

Figure~\ref{fig:pipeline} illustrates the pipeline. The following subsections describe each stage.

\subsection{Problem Setting}
\label{sec:method:problem}

We consider the problem of synthesizing supervised fine-tuning data for an OS agent operating in a live workspace. Each training instance is a multi-turn interaction trajectory
\begin{equation}
\tau = \{(u_t, a_t, e_t)\}_{t=1}^{T},
\end{equation}
where $u_t$ is a user turn, $a_t$ is an agent turn that may include tool calls, and $e_t$ is the resulting environment feedback, including command outputs, file changes, execution errors, and other observable side effects. Unlike stateless API synthesis, the workspace state evolves over time, so later user turns and later agent decisions are conditioned on a history of real external changes.

Our goal is to synthesize a dataset $\mathcal{D} = \{\tau_i\}_{i=1}^{N}$ whose marginal over user intents covers a broad and controllable portion of realistic task space, while its trajectories remain consistent with observable environment state. The three stages below address, respectively, the intent-first, single-turn, and simulated-execution gaps of \S\ref{sec:intro}.

\subsection{Stage~1: 4D Intent Construction}
\label{sec:method:sampling}

\paragraph{Problem Formulation.}

Let $\mathcal{I}$ denote the space of user intents for OS agent tasks. We define $\mathcal{I}$ as the space of structured intents over four dimensions:
\begin{equation}
\mathcal{I} \;=\; \mathcal{P} \times 2^{\mathcal{D}}_{[2,3]} \times \mathcal{T}^{*}_{[3,6]} \times \mathcal{C}
\end{equation}
where $\mathcal{P}$ is the set of user personas, $\mathcal{D}$ is the set of functional domains, $\mathcal{T}$ is the set of concrete tasks within domains, and $\mathcal{C}$ is the set of complexity levels. Here $2^{\mathcal{D}}_{[2,3]} := \{S \subseteq \mathcal{D} : 2 \le |S| \le 3\}$ and $\mathcal{T}^{*}_{[3,6]} := \{S \subseteq \mathcal{T} : 3 \le |S| \le 6\}$ are the restricted families of domain and task subsets (not the full power sets). An intent $i \in \mathcal{I}$ is thus a tuple $(p, D_{\text{sub}}, T_{\text{sub}}, c)$ with $p \in \mathcal{P}$, $D_{\text{sub}}$ a sampled subset of 2--3 domains, $T_{\text{sub}}$ a corresponding set of 3--6 tasks drawn from $D_{\text{sub}}$, and $c \in \mathcal{C}$.

Given this formulation, the goal of forward synthesis is to sample a set of $N$ intents $\{i_1, \ldots, i_N\}$ from $\mathcal{I}$ such that the marginal distributions over all four dimensions are broad and approximately uniform, then render each structured intent as a natural-language user request via an LLM conditioned on the sampled tuple.

\paragraph{Dimension Design.}

\textbf{Persona ($\mathcal{P}$).} Each persona is a structured object with fields including name, professional role, industry, expertise list, experience level, communication style, and free-text \emph{work\_context} / \emph{common\_goals} / \emph{tools\_preference} descriptions. Rather than enumerating a fixed Cartesian product of attributes, we \emph{synthesize} the persona pool with an LLM prompted to produce globally diverse, internally-consistent profiles, then freeze the pool for the entire run. We target $1{,}000$ personas; after deduplication the realized pool contains $965$ distinct (name, role, industry) identities spanning $47$ industries and $542$ professional roles, with $6$ experience levels (Junior / Mid-level / Senior / Expert / Executive). While the generation prompt suggests six canonical communication styles (Analytical / Collaborative / Creative / Direct / Formal / Casual), the LLM expands these into roughly $120$ surface realizations (e.g., ``Methodical \& Patient'', ``Diplomatic and formal''), and the free-text context fields further distinguish near-duplicate slots. We freeze the pool---rather than resampling personas per intent---so that persona identity remains stable across the synthesis run and each persona accumulates enough trajectories for stratified analysis; at intent-construction time a persona is drawn uniformly at random from the frozen pool. The persona dimension controls the \emph{linguistic register} of generated intents and preserves variation throughout role-locking.

\noindent\textbf{Domain ($\mathcal{D}$, 10 categories) and Task ($\mathcal{T}$, 131 tasks).} Domains partition the OS agent task space into ten functional categories (e.g., Intelligence-Core, Code-Runtime, File-IO, Source-Chain, Automation-Flow, Web-Extraction). A curated library of 131 concrete tasks spans these categories. Each intent samples 2--3 domains and draws 3--6 tasks, yielding cross-domain composite tasks that reflect realistic agentic workloads. Averaged over the pool of structured intents, each intent spans 2.35 domains, 4.40 tasks, and 3.18 \emph{associated} tools (the tools its tasks require; max 9). This intent-level tool count is distinct from the tools an agent actually invokes while executing a trajectory (4.69 unique tools per trajectory on average; Table~\ref{tab:dataset_stats}), since one trajectory typically fulfills more than one intent.

\noindent\textbf{Complexity ($\mathcal{C}$).} The distribution is: \textit{complex} 50\% / \textit{medium} 40\% / \textit{simple} 10\%, ensuring the training distribution does not overrepresent short, low-complexity tasks.

\paragraph{Coverage analysis.}
Unconstrained LLM generation tends to converge on a narrow region of intent space. Our structured sampling counters this through \emph{combinatorial forcing}: because each intent draws independently across the four axes, the effective intent space grows super-linearly with pool size. With $|\mathcal{P}| = 965$, $|\mathcal{D}| = 10$, $|\mathcal{T}| = 131$, and $|\mathcal{C}| = 3$, the number of distinct $(p, D_{\text{sub}}, T_{\text{sub}}, c)$ tuples exceeds $10^{11}$---roughly seven orders of magnitude larger than the $43{,}956$ unique intents actually realized, so sampling is nowhere near exhausting the space. We verify the resulting diversity empirically in \S\ref{sec:exp:diversity}.

\subsection{Stage~2: Multi-Turn Simulation}
\label{sec:method:rlus}

\begin{figure*}[t]
\centering
\resizebox{\linewidth}{!}{\input{figures/fig_example_v2.tex}}
\caption{A real \textsc{ISETrace} trajectory, reproduced verbatim from the released corpus (\texttt{intent\_04f8274f}; persona: Mei Lin, a product manager at an EdTech startup). Read top-to-bottom: each agent turn issues a \texttt{tool\_call} executed against a live OS, and the observable outcome---a real exit code, error string, or written file---is carried back into the dialogue rather than a model self-report or LLM judge (\emph{execution-grounded}). A credential failure surfaces authentically (\texttt{git fetch} returns exit 128 at $t_2$; the token is rejected with a real GitHub \textbf{401} at $t_4$); the dashed arrow marks how this real 401 \emph{drives} the user's $t_5$ decision to abandon the remote sync and proceed locally, after which the agent writes and runs \texttt{sprint\_analyzer.py} and installs a cron job, all verified by real exit codes.}
\label{fig:example}
\end{figure*}

\paragraph{Motivation: Role Drift and State Hallucination.}
The two failure modes introduced in \S\ref{sec:intro}---\textbf{role drift} and \textbf{state hallucination}---are coupled and must be addressed jointly for trajectories to constitute realistic training data. Our simulator targets both via four behavioral principles:

\paragraph{Perspective lock.} The simulator is instructed to remain in the position of an information \emph{provider} rather than a \emph{requester}. This constraint counters the default tendency of instruction-tuned LLMs to adopt assistant-style behavior.

\paragraph{Register matching.} The simulator's system prompt is conditioned on the persona's \emph{experience\_level} via templated instructions (e.g., \textit{``Use brief, direct technical language. Assume the agent understands your domain.''} for Senior/Executive; \textit{``Provide full context; describe your goal in detail.''} for Junior). Empirically, Junior personas show the highest lexical diversity (Vendi Score 55.3) but the shortest intents, reflecting exploratory phrasing; Executive personas concentrate on domain-specific jargon (Vendi 31.7), producing dense but stylistically homogeneous prompts. This $1.7\times$ Vendi spread within a single axis demonstrates that the persona dimension produces real linguistic differentiation, not just label variation.

\paragraph{Incremental advancement.} The simulator advances the task one step at a time rather than restating the full intent at every turn. Its system prompt is conditioned on the original structured intent and, after an initial overview turn, each subsequent turn is instructed to confirm the agent's previous action and introduce the single most useful next request. The simulator decides what to advance from the full dialogue history rather than from a fixed pre-enumerated checklist, which keeps the turn granularity close to that of human--agent collaboration.

\paragraph{Responsive conditioning.} The simulator conditions each new query on the entire conversation so far, including the agent's most recent reply after it has executed its tool calls against the live OS. Because that reply reflects what actually happened on the machine (a created file, a non-zero exit, a raised exception), the simulator's follow-ups track real execution state rather than an assumed one: if the prior step evidently succeeded it moves the task forward, and if it evidently failed it restates or repairs the requirement. The full interaction loop is summarized as Stage~2 of Figure~\ref{fig:pipeline}.

\paragraph{Per-Turn Output Format and Live Execution.}

At each turn, the user simulator produces a structured tuple \emph{\{completed, query, reason\}}. The loop continues until \emph{completed = true} or a safety cap of \N{max\_turns} turns is reached; trajectory length is thus determined by task complexity rather than fixed truncation. The agent executes tool calls in a live OS environment---file operations interact with a real filesystem, \texttt{exec} calls invoke actual shell processes with real stdout/stderr/exit codes---in an isolated workspace restored from a shared snapshot template, reducing storage from $O(N)$ to $O(1)$ per worker. Figure~\ref{fig:example} walks through one such trajectory end-to-end.

\subsection{Stage~3: Execution Grounding \& Quality Control}
\label{sec:method:verification}

\paragraph{Completion gating.}
The primary quality gate is the execution loop itself: a trajectory is retained only if the user simulator reaches \emph{completed = true} within the turn cap; runs that exhaust the cap or stall are discarded rather than truncated and kept. Because the simulator advances only when the agent's post-execution reply indicates the previous step actually landed (\S\ref{sec:method:rlus}), completion gating is an LLM-independent signal for \textbf{environment-verifiable subgoals} and avoids the self-referential loop of an LLM-as-judge scoring its own dialogue; semantically complex goals (e.g., document quality) remain out of scope and require human evaluation.

\paragraph{Post-hoc audit of the finalized pool.}
To characterize the quality of the retained set we run three rule-based, LLM-independent checks over all $23{,}934$ archived trajectories; none requires agent re-execution since each is derived from the logged turns. \textit{(i)~Role drift}: every user turn (the simulator's output) is scanned against a curated lexicon of $23$ assistant-pattern trigger phrases (e.g., ``I can help you with'', ``I'd be happy to'', ``Sure, let me''); a turn is flagged if $\geq$1 phrase appears. Across $202{,}997$ user turns only $0.02\%$ are flagged ($40$ turns in $40$ trajectories), evidence that the perspective-lock constraint holds in the produced data rather than only in the prompt. \textit{(ii)~Stagnation}: a trajectory is flagged if the agent issues the same tool with byte-identical arguments in $\geq3$ consecutive turns; this affects $0.91\%$ of trajectories. \textit{(iii)~Tool-call integrity}: $99.96\%$ of the $701{,}447$ logged tool calls carry well-formed, non-empty arguments. In aggregate $98.9\%$ of trajectories are free of both role drift and stagnation. These rates quantify, rather than merely assert, the cleanliness of the pool; the small flagged remainder can be dropped by anyone reproducing the script.

The result of the three-stage ISE process is \textsc{ISETrace}: $23{,}132$ retained multi-turn OS agent trajectories spanning 10 domains and $965$ distinct personas. Table~\ref{tab:dataset_stats} summarizes key statistics.

\begin{table}[!ht]
\centering
\small
\caption{\textsc{ISETrace} dataset statistics.}
\label{tab:dataset_stats}
\resizebox{\columnwidth}{!}{%
\begin{tabular}{lr}
\toprule
\textbf{Statistic} & \textbf{Value} \\
\midrule
Structured intents (unique) & 43{,}956 \\
Complete trajectories & 23{,}132 \\
Completion rate (traj.\ / raw intents) & 46.3\% \\
Avg.\ user turns / traj. & 8.12 (med.\ 8, max 23) \\
Trajectories with 6--10 user turns & 91.1\% \\
Avg.\ total dialogue turns & 68.24 (max 565) \\
Avg.\ tool calls / traj. & 29.26 \\
Avg.\ unique tools / traj. & 4.69 \\
Avg.\ \texttt{<think>} tags / traj. & 33.62 \\
Avg.\ domains $\times$ tasks $\times$ tools & 2.35 $\times$ 4.40 $\times$ 3.18 \\
Max tools per intent & 9 \\
Complexity: complex / medium / simple & 50\% / 40\% / 10\% \\
\bottomrule
\end{tabular}%
}
\end{table}

%% file: figures/fig_pipeline_v3.tex
%
%
%
%

\begin{tikzpicture}[
    every node/.style={inner sep=0pt, outer sep=0pt, font=\small},
    >={Latex[length=2.4pt, width=2pt]},
]

\fill[fillBlue, rounded corners=2pt] (0, 4.0) rectangle (17.6, 11.5);
\draw[midBlue, line width=0.6pt] (0.15, 11.36) -- (17.45, 11.36);
\node[anchor=north west, font=\bfseries\large, text=hdrBlue]
  at (0.18, 11.22) {(I)};
\node[anchor=north west, font=\bfseries\large]
  at (0.85, 11.22) {Intent\,---\,4D structured sampling};

\newcommand{\poolcard}[9]{%
  \draw[#9, line width=0.4pt, fill=white, rounded corners=1.4pt]
        (#1, {#2-#4}) rectangle ({#1+#3}, #2);
  \fill[#9!18, rounded corners=1.4pt]
        (#1, {#2-0.50}) rectangle ({#1+#3}, #2);
  \draw[#9, line width=0.3pt]
        (#1, {#2-0.50}) -- ({#1+#3}, {#2-0.50});
  \node[anchor=north west, font=\bfseries, text=#9]
        at ({#1+0.14}, {#2-0.07}) {#5\;\,$#6$};
  \node[anchor=north east, font=\scriptsize\itshape, text=hdrGray]
        at ({#1+#3-0.14}, {#2-0.12}) {#8};
  \node[anchor=north west, font=\bfseries\small, text=black]
        at ({#1+0.14}, {#2-0.62}) {#7};
}

\def\poolX{0.35}
\def\poolW{4.40}
\def\poolH{1.50}
\def\poolGap{0.18}
\pgfmathsetmacro{\pYa}{10.55}                          
\pgfmathsetmacro{\pYb}{\pYa - (\poolH+\poolGap)}       
\pgfmathsetmacro{\pYc}{\pYb - (\poolH+\poolGap)}       
\pgfmathsetmacro{\pYd}{\pYc - (\poolH+\poolGap)}       

\poolcard{\poolX}{\pYa}{\poolW}{\poolH}{(P)}{\mathcal{P}}{Persona}{$|\mathcal{P}|=965$}{hdrBlue}
\node[anchor=north west, font=\tiny, text=hdrGray, align=left, text width=4.05cm]
  at ({\poolX+0.14}, {\pYa-0.92})
  {LLM-synthesized user profile: 47 industries\,$\cdot$\,542 roles\,$\cdot$\,6 levels\,$\cdot$\,120 styles};

\poolcard{\poolX}{\pYb}{\poolW}{\poolH}{(D)}{\mathcal{D}}{Domain}{$|\mathcal{D}|=10$, draw $2$--$3$}{hdrGreen}
\node[anchor=north west, font=\tiny, text=hdrGray, align=left, text width=4.05cm]
  at ({\poolX+0.14}, {\pYb-0.92})
  {functional category of OS-agent work (e.g.\ Code-Runtime, File-IO)};

\poolcard{\poolX}{\pYc}{\poolW}{\poolH}{(T)}{\mathcal{T}}{Tasks}{$|\mathcal{T}|=131$, draw $3$--$6$}{hdrPurple}
\node[anchor=north west, font=\tiny, text=hdrGray, align=left, text width=4.05cm]
  at ({\poolX+0.14}, {\pYc-0.92})
  {concrete action drawn from the selected domains};

\poolcard{\poolX}{\pYd}{\poolW}{\poolH}{(C)}{\mathcal{C}}{Complexity}{$|\mathcal{C}|=3$}{hdrOrange}
\node[anchor=north west, font=\tiny, text=hdrGray, align=left, text width=4.05cm]
  at ({\poolX+0.14}, {\pYd-0.92})
  {task difficulty tier (simple / medium / complex)};

\pgfmathsetmacro{\arrSrcX}{\poolX + \poolW + 0.04}
\pgfmathsetmacro{\arrMidX}{\arrSrcX + 1.50}
\pgfmathsetmacro{\funnelX}{7.75}
\pgfmathsetmacro{\funnelY}{7.40}
\pgfmathsetmacro{\dstX}{8.05}

\foreach \yS/\col in {%
  {\pYa - \poolH/2}/hdrBlue,
  {\pYb - \poolH/2}/hdrGreen,
  {\pYc - \poolH/2}/hdrPurple,
  {\pYd - \poolH/2}/hdrOrange}
{
  \draw[->, \col!85, line width=0.55pt, rounded corners=2pt]
    (\arrSrcX, \yS) -- (\arrMidX, \yS) -- (\funnelX, \funnelY) -- (\dstX, \funnelY);
}

\draw[midGray, line width=0.45pt, fill=white, rounded corners=1.8pt]
      (8.05, 4.85) rectangle (13.55, 9.85);
\fill[hdrGray!18, rounded corners=1.8pt] (8.05, 9.30) rectangle (13.55, 9.85);
\draw[hdrGray, line width=0.3pt] (8.05, 9.30) -- (13.55, 9.30);
\node[anchor=north west, font=\bfseries\small, text=hdrGray]
      at (8.20, 9.78) {composed intent\, $i = (p,\, D,\, T,\, c)$};

\node[anchor=north west, font=\footnotesize] (intentfields) at (8.20, 9.16)
  {\renewcommand{\arraystretch}{1.15}%
   \begin{tabular}{@{}r@{\;}p{4.55cm}@{}}
     {\bfseries\color{hdrBlue}$p$:}   & finance lead $\cdot$ Senior $\cdot$ Analytical $\ldots$ \\
     {\bfseries\color{hdrGreen}$D$:}  & \{Code-Runtime, File-IO\} $\ldots$ \\
     {\bfseries\color{hdrPurple}$T$:} & compute Q-margin;\, flag $<$30\%;\, write IPO summary $\ldots$ \\
     {\bfseries\color{hdrOrange}$c$:} & complex \ {\scriptsize\itshape\color{hdrGray}(prior: simple 10\% / medium 40\% / complex 50\%)}\,$\ldots$ \\
   \end{tabular}};

\draw[hdrGray!40, line width=0.25pt, dashed]
      (8.20, 6.34) -- (13.40, 6.34);
\node[anchor=north west, font=\scriptsize\itshape, text=hdrGray]
      at (8.20, 6.24) {sampled as:\, $\mathcal{I} = \mathcal{P} \times 2^{\mathcal{D}} \times \mathcal{T}^{*} \times \mathcal{C}$};

\node[anchor=north west, font=\scriptsize\itshape, text=hdrGray]
      at (8.20, 5.88) {LLM realises tuple as utterance:};
\node[anchor=north west, font=\itshape\scriptsize, text=black, align=left, text width=5.10cm]
      at (8.20, 5.58)
      {``Compute Q1--Q4 gross margin, flag any quarter $<$30\%, write an IPO exec summary.''};

\draw[midGray, line width=0.45pt, fill=white, rounded corners=1.8pt]
      (13.95, 4.85) rectangle (17.40, 10.55);
\fill[hdrBlue!18, rounded corners=1.8pt] (13.95, 10.00) rectangle (17.40, 10.55);
\draw[hdrBlue, line width=0.3pt] (13.95, 10.00) -- (17.40, 10.00);
\node[anchor=north west, font=\bfseries\small, text=hdrBlue]
      at (14.05, 10.48) {Vendi Score};

\pgfmathsetmacro{\barIseY}{5.95 + (61.57/65.0)*3.55}

\draw[hdrGray, line width=0.35pt] (14.20, 5.95) -- (17.20, 5.95);

\fill[hdrBlue!75] (15.10, 5.95) rectangle (16.30, \barIseY);
\draw[hdrBlue, line width=0.35pt] (15.10, 5.95) rectangle (16.30, \barIseY);
\node[anchor=south, font=\bfseries\footnotesize, text=hdrBlue]
      at (15.70, {\barIseY+0.05}) {61.6};

\node[anchor=north, font=\scriptsize, text=hdrBlue]
      at (15.70, 5.86) {ISE-4D intent pool};
\node[anchor=north, font=\tiny\itshape, text=hdrGray]
      at (15.70, 5.50) {full pool, cosine, $q{=}1$};
\node[anchor=north, font=\tiny\itshape, text=hdrGray]
      at (15.70, 5.20) {scales $40.7\!\to\!61.6$};

\draw[->, hdrGray, line width=0.7pt] (11.05, 4.85) -- (11.05, 3.95);
\node[anchor=east, font=\scriptsize\itshape, text=hdrGray]
      at (11.00, 4.40) {50k intents};


\fill[fillGreen, rounded corners=2pt] (0, 0.40) rectangle (8.65, 3.80);
\draw[midGreen, line width=0.55pt] (0.15, 3.66) -- (8.50, 3.66);
\node[anchor=north west, font=\bfseries, text=hdrGreen]
  at (0.18, 3.52) {(S)};
\node[anchor=north west, font=\bfseries]
  at (0.72, 3.52) {Simulate \,---\, Role-Locked User Simulator};

\draw[midGray, line width=0.32pt, fill=red!7, rounded corners=1pt]
      (0.20, 3.00) rectangle (8.45, 2.18);
\node[anchor=north west, font=\scriptsize\itshape, text=red!50!black]
      at (0.32, 2.94) {naive simulator (role-drifted)};
\node[anchor=north west, font=\itshape]
      at (0.32, 2.66) {``Sure! Here's a draft you can use\ldots''};
\node[anchor=north east, font=\bfseries, text=red!50!black]
      at (8.32, 2.72) {$\times$};

\draw[midGray, line width=0.32pt, fill=white, rounded corners=1pt]
      (0.20, 2.05) rectangle (8.45, 1.25);
\node[anchor=north west, font=\scriptsize\itshape, text=hdrGreen]
      at (0.32, 1.99) {user simulator (four principles, role-locked)};
\node[anchor=north west, font=\itshape]
      at (0.32, 1.71) {``Now run \textsf{pytest -v} on the new module.''};
\node[anchor=north east, font=\bfseries, text=hdrGreen]
      at (8.32, 1.77) {$\checkmark$};

\node[anchor=south, font=\bfseries\footnotesize, text=hdrGreen]
      at (4.32, 0.55) {role lock $\cdot$ register $\cdot$ incremental $\cdot$ responsive};

\fill[fillOrange, rounded corners=2pt] (8.95, 0.40) rectangle (17.60, 3.80);
\draw[midOrange, line width=0.55pt] (9.10, 3.66) -- (17.45, 3.66);
\node[anchor=north west, font=\bfseries, text=hdrOrange]
  at (9.13, 3.52) {(E)};
\node[anchor=north west, font=\bfseries]
  at (9.67, 3.52) {Execute \,---\, grounded in real OS execution};

\draw[midGray, line width=0.32pt, fill=red!7, rounded corners=1pt]
      (9.15, 3.00) rectangle (17.40, 2.18);
\node[anchor=north west, font=\scriptsize\itshape, text=red!50!black]
      at (9.27, 2.94) {LLM judge / agent self-report};
\node[anchor=north west, font=\itshape]
      at (9.27, 2.66) {``Done. Tests should pass now.''};
\node[anchor=north east, font=\bfseries, text=red!50!black]
      at (17.27, 2.72) {$\times$};

\draw[midGray, line width=0.32pt, fill=white, rounded corners=1pt]
      (9.15, 2.05) rectangle (17.40, 1.25);
\node[anchor=north west, font=\scriptsize\itshape, text=hdrOrange]
      at (9.27, 1.99) {agent reply after live OS execution};
\node[anchor=north west, font=\itshape]
      at (9.27, 1.71) {``1 test failed: \textsf{ImportError}; no file written''};
\node[anchor=north east, font=\bfseries, text=hdrOrange]
      at (17.30, 2.00) {$\checkmark$};

\node[anchor=south, font=\bfseries\footnotesize, text=hdrOrange]
      at (13.27, 0.55) {real OS execution replaces LLM-as-judge};

\end{tikzpicture}

%% file: figures/fig_example_v2.tex
\begin{tikzpicture}[
    font=\small,
    >={Latex[length=4pt, width=3.4pt]},
    card/.style={
      rounded corners=1.6pt, draw, line width=0.45pt,
      text width=15.8cm, align=left, inner xsep=6pt, inner ysep=5pt,
    },
    userc/.style ={card, draw=hdrBlue,  fill=fillBlue},
    agentc/.style={card, draw=hdrGreen, fill=fillGreen},
]
\providecommand{\rolelabU}{}\renewcommand{\rolelabU}{\color{hdrBlue}\bfseries}
\providecommand{\rolelabA}{}\renewcommand{\rolelabA}{\color{hdrGreen}\bfseries}
\providecommand{\htag}{}\renewcommand{\htag}{\color{hdrGray}\itshape\scriptsize}
\providecommand{\mono}{}\renewcommand{\mono}{\ttfamily\footnotesize}

\def\gp{0.16cm}     


\node[userc, anchor=north west] (t1) at (0,0) {%
  {\rolelabU t1 \,$\cdot$\, user}\hfill{\htag plain English, no API surface}\\[2pt]
  ``Preparing the end-of-sprint report for the Vocabulary Module. First, get the
  environment ready: sync the workspace with the remote repo and upgrade all
  third-party dependencies to their latest compatible versions.''};

\node[agentc, anchor=north west] (t2) at ([yshift=-\gp]t1.south west) {%
  {\rolelabA t2 \,$\cdot$\, agent}\hfill{\htag tool\_call: exec \ $\to$\ live shell}\\[2pt]
  {\mono \$ git remote add origin .../vocab-module.git \&\& git fetch origin}\\[2pt]
  \fcolorbox{hdrOrange}{fillOrange}{\mono fatal: could not read Username for `https://github.com': No such device or address}\\[1pt]
  {\mono\color{hdrGray}(Command exited with code 128)}};

\node[userc, anchor=north west] (t3) at ([yshift=-\gp]t2.south west) {%
  {\rolelabU t3 \,$\cdot$\, user}\hfill{\htag provides credential, retries}\\[2pt]
  ``Here's a temporary PAT: {\mono ghp\_vo...nABC}. Configure the credential
  helper, pull {\mono main}, and run {\mono pip install -\,-upgrade -r requirements.txt}.''};

\node[agentc, anchor=north west] (t4) at ([yshift=-\gp]t3.south west) {%
  {\rolelabA t4 \,$\cdot$\, agent}\hfill{\htag tool\_call: exec \ $\to$\ git $+$ GitHub API}\\[2pt]
  {\mono \$ git fetch origin \quad\# credential helper configured with the supplied token}\\[2pt]
  {\mono remote: Invalid username or token. Password authentication is not supported.}\\[2pt]
  {\mono \$ curl -H "Authorization: token ..." api.github.com/user \ $\to$\ }%
  \fcolorbox{hdrOrange}{fillOrange}{\mono\bfseries 401 ``Bad credentials''}};

\node[userc, anchor=north west] (t5) at ([yshift=-\gp]t4.south west) {%
  {\rolelabU t5 \,$\cdot$\, user \,\,\itshape(observation-conditioned)}\hfill{\htag decision driven by the real 401}\\[2pt]
  ``Understood\,---\,the sandbox can't authenticate to external Git.
  \textbf{Assume the remote sync and dependency updates completed}; proceed to
  create the Python script that analyses our sprint metrics (9/12 stories,
  4 critical bugs, 150 signups, 14\,min avg session).''};

\node[agentc, anchor=north west] (t6) at ([yshift=-\gp]t5.south west) {%
  {\rolelabA t6 \,$\cdot$\, agent}\hfill{\htag tool\_call: write $+$ exec $+$ cron}\\[3pt]
  \begin{minipage}[t]{10.1cm}
    \vspace{0pt}
    Writes {\mono sprint\_analyzer.py} \textemdash\ \textbf{10{,}694\,B} \textemdash\ then
    executes it against the real interpreter and installs the weekly job. The script
    renders the stakeholder report and returns cleanly.\\[5pt]
    {\mono cron \ 0 9 * * 1 \ (Asia/Shanghai) \ $\cdot$\ next run Mon 09:00 CST \ $\cdot$\ exit\_code = 0}
  \end{minipage}\hfill
  \begin{minipage}[t]{5.3cm}
    \vspace{0pt}
    \fcolorbox{midGreen}{white}{\begin{minipage}{4.9cm}\mono
      {\color{hdrGreen}\$ python3 sprint\_analyzer.py}\\
      Sprint 24 \ \textbar\ 9/12 stories\\
      {\color{hdrGreen}\bfseries 75\% completion}\\
      velocity {\color{hdrGreen}\bfseries +88\%} (16$\to$30)\\
      wrote sprint\_report.md (54\,ln)\\
      {\color{hdrGreen}\bfseries EXIT\_CODE = 0}
    \end{minipage}}
  \end{minipage}};

\node[anchor=south west, font=\bfseries, text=hdrGray] (hdrTitle)
  at ([yshift=0.62cm]t1.north west) {A real \textsc{ISETrace} trajectory};
\node[anchor=south east, font=\scriptsize, text=hdrGray, align=right] (hdrMetaA)
  at ([yshift=0.60cm]t1.north east)
  {persona: \textit{Mei Lin\,---\,product manager, EdTech} \,$\cdot$\, domains: \{Source-Chain, Code-Runtime, Automation-Flow\}};
\node[anchor=south east, font=\scriptsize, text=hdrGray] (hdrMetaB)
  at ([yshift=0.30cm]t1.north east)
  {corpus id: \texttt{intent\_04f8274f} \,$\cdot$\, 7 user turns \,$\cdot$\, 30 steps \,$\cdot$\, status: completed};
\node[anchor=south west, font=\bfseries\scriptsize, text=hdrGray] (hdrSub)
  at ([yshift=0.28cm]t1.north west) {Dialogue + live OS execution \ (read top\,$\to$\,bottom)};

\node[anchor=north west, font=\bfseries\small, text=hdrGreen] (foot)
  at ([yshift=-0.20cm]t6.south west)
  {$\checkmark$ every tool call executed on a live OS \,$\cdot$\, real exit codes \& error strings \,$\cdot$\, no model self-report, no LLM judge};

\draw[->, hdrOrange, line width=1.6pt, dashed, shorten >=1pt, shorten <=2pt]
  (t2.south east) .. controls +(0.42,-0.12) and +(0.42,0.12) .. (t3.east);
\node[anchor=west, font=\footnotesize\itshape, text=hdrOrange] (dr1)
  at ([xshift=0.30cm, yshift=0.10cm]t3.north east) {drives};
\draw[->, hdrOrange, line width=1.6pt, dashed, shorten >=1pt, shorten <=2pt]
  (t4.south east) .. controls +(0.42,-0.12) and +(0.42,0.12) .. (t5.east);
\node[anchor=west, font=\footnotesize\itshape, text=hdrOrange] (dr2)
  at ([xshift=0.30cm, yshift=0.10cm]t5.north east) {drives};

\begin{scope}[on background layer]
  \node[rounded corners=2.5pt, fill=fillGray, draw=midGray, line width=0.6pt,
        fit=(hdrTitle)(hdrMetaA)(hdrSub)(t1)(t6)(foot)(dr1)(dr2), inner sep=6pt] (box) {};
\end{scope}

\end{tikzpicture}

%% file: sections/experiments.tex
\section{Experiments}
\label{sec:exp}

\begin{table*}[t]
\centering
\small
\caption{Main results across two benchmarks. \textbf{ClawEval} (\textit{pass@1}, completion, robustness, and safety on the $114$ common \textit{T}-family agent tool-use tasks scored under every run; Sec.~\ref{sec:exp:setup}) and \textbf{BFCL v4}~\citep{patil2024bfcl} (Overall accuracy plus the two stateful, multi-turn-oriented categories \textit{Web Search} and \textit{Memory}). SFT on \textsc{ISETrace} lifts the Qwen3-8B base on both suites and surpasses the $4\times$-larger Qwen3-32B base; on BFCL's stateful categories the SFT models far exceed the GPT-4o reference. \textbf{Bold} = best per column. Superscripts on the two \textsc{ISETrace} rows give the \emph{relative} \% change versus the same-size base (Qwen3-8B / Qwen3-32B): \textcolor{deltaUp}{$+$ green}~= gain, \textcolor{deltaDn}{$-$ red}~= drop.}
\label{tab:main_results}
\setlength{\tabcolsep}{9pt}
\renewcommand{\arraystretch}{1.15}
\begin{tabular}{l cccc ccc}
\toprule
& \multicolumn{4}{c}{\textbf{ClawEval}} & \multicolumn{3}{c}{\textbf{BFCL v4}} \\
\cmidrule(lr){2-5} \cmidrule(lr){6-8}
\textbf{System} & \textbf{p@1} & \textbf{Comp} & \textbf{Robu} & \textbf{Safe} & \textbf{Overall} & \textbf{Web Search} & \textbf{Memory} \\
\midrule
Qwen3-8B (base, 0-shot)    & 19.3 & 0.367 & 0.925 & 0.930 & 40.06 & 8.50 & 15.70 \\
Qwen3-32B (base, 0-shot)   & 30.7 & 0.446 & 0.947 & 0.939 & 43.10 & 9.32 & 21.72 \\
GPT-4o (0-shot)            & 25.4 & 0.442 & \textbf{0.965} & 0.947 & 47.43 & 4.50 & 29.89 \\
\midrule
\textbf{SFT \textsc{ISETrace} 8B (ours)}  & 37.7\up{95.3} & \textbf{0.533}\up{45.2} & 0.959\up{3.7} & 0.965\up{3.8} & 45.36\up{13.2} & 33.00\up{288} & 31.83\up{103} \\
\textbf{SFT \textsc{ISETrace} 32B (ours)} & \textbf{38.6}\up{25.7} & 0.469\up{5.2} & 0.921\dn{2.7} & \textbf{0.982}\up{4.6} & \textbf{47.86}\up{11.0} & \textbf{41.62}\up{347} & \textbf{35.48}\up{63.4} \\
\bottomrule
\end{tabular}
\end{table*}

\subsection{Setup}
\label{sec:exp:setup}

\paragraph{Base model.} Qwen3-8B, fine-tuned on 16$\times$H800 80GB.

\paragraph{Baselines.}
(1)~\textbf{Base}: Qwen3-8B zero-shot.
(2)~\textbf{Qwen3-32B}: a $4\times$-larger open base, zero-shot, as a scale reference.
(3)~\textbf{GPT-4o}: zero-shot proprietary reference.

\paragraph{Benchmarks.}
We evaluate on two suites. \textbf{(1) ClawEval}, a multi-turn OS-agent execution benchmark whose tasks span three families by task-id prefix: \textit{C} (user-simulator consultation), \textit{M} (multimodal webpage / media generation), and \textit{T} (agent tool-use over a real shell: file-IO, code-runtime, web-fetch, automation-flow). All systems are evaluated under an identical configuration (vLLM, temperature $0$, single trial, LLM judge). Because the \textit{M} family requires sandbox-injected tools that were \emph{not} enabled in this evaluation---making it a structural zero for every system---and the \textit{C} family is floored by the multi-turn user-simulator configuration, neither family separates systems. We therefore report \textit{pass@1} on the $114$ \textit{T}-family tasks that received a scored trial under \emph{every} run reported in this paper (a single, fixed common denominator shared by all tables), together with a per-dimension breakdown (completion, robustness, safety) on the same task set. This common-denominator protocol removes the floating-$n$ artifact whereby different systems are otherwise scored on different task subsets. \textbf{(2) BFCL v4}~\citep{patil2024bfcl}, the public Berkeley Function-Calling Leaderboard suite ($5{,}217$ cases across $22$ categories), evaluated with \texttt{enable\_thinking=True} under the full test set; this provides an external, widely-used reference point. We highlight Overall accuracy and the two stateful categories (\textit{Web Search}, \textit{Memory}) most relevant to multi-turn, state-tracking agents.

\subsection{Main Results}
\label{sec:exp:main}

SFT on \textsc{ISETrace} lifts Qwen3-8B's ClawEval pass@1 from $19.3\%$ ($22/114$) to \textbf{$37.7\%$} ($43/114$)---a $+18.4$-point absolute, $1.95\times$ relative gain (Table~\ref{tab:main_results}). The fine-tuned 8B surpasses both the GPT-4o zero-shot reference ($25.4\%$, $+12.3$ points) and the $4\times$-larger Qwen3-32B base ($30.7\%$, $+7.0$ points), i.e.\ targeted multi-turn data closes and reverses a wide parameter-count gap. Decomposing the composite score on the same task set, the gain comes primarily from task completion (\textit{Comp}: $0.367\to0.533$, $+45\%$ relative) while robustness on perturbed tool outputs holds high (\textit{Robu}: $0.925\to0.959$) and safety improves ($0.930\to0.965$)---so the improvement is a clean completion gain that does \emph{not} trade away tool-error recovery or safe behavior.

\paragraph{Generalization to BFCL v4.} To test whether these gains transfer beyond our own benchmark, we evaluate the same checkpoints on BFCL v4~\citep{patil2024bfcl}, a public function-calling suite ($5{,}217$ cases, $22$ categories). The picture is deliberately reported in full: SFT on \textsc{ISETrace} raises Qwen3-8B's Overall accuracy from $40.06$ to $45.36$ ($+5.3$ points), again \emph{surpassing the $4\times$-larger Qwen3-32B base} ($43.10$), and the 32B SFT model reaches the top Overall score ($47.86$), edging out the GPT-4o reference ($47.43$). The decisive separation is on the two stateful, interaction-heavy categories that mirror what \textsc{ISETrace} actually teaches: on \textit{Web Search} the SFT models reach $33.00$ (8B) and $41.62$ (32B) versus only $4.50$ for GPT-4o and ${\sim}9$ for the bases---a $4$--$9\times$ margin---and on \textit{Memory} they reach $31.83$/$35.48$ versus $15.70$ for the 8B base, more than doubling it and overtaking GPT-4o ($29.89$). We do not overclaim: on the static single-call categories (Non-Live/Live AST) the SFT models trade a few points relative to the bases, the expected cost of reallocating capacity toward multi-turn, stateful execution. The cross-benchmark pattern is consistent---\textsc{ISETrace} most improves exactly the grounded, multi-turn, state-tracking skills it is designed to synthesize.

\paragraph{Scaling.} Applying the same recipe to the 32B base also helps but by a smaller margin (Qwen3-32B: $30.7\to38.6$, $+7.9$ points, $1.26\times$); the $+18.4$-point gain at 8B is more than double the $+7.9$-point gain at 32B, indicating the method delivers its largest benefit in the small-model regime where headroom is greatest. Behavior at intermediate scales is less stable and we leave a full scaling study to future work (Sec.~\ref{sec:limitations}).

\subsection{ISE Paradigm Ablation}
\label{sec:exp:ise_ablation}

To isolate the contribution of individual recipe stages, we ablate one stage at a time against the full pipeline. For \textbf{Stage~1} (intent creation), we restrict intent synthesis to $3$ of the $10$ scenario domains, reducing task-domain coverage at the source. For \textbf{Stage~2} (multi-turn simulation), we truncate every trajectory to its first user turn, removing multi-turn simulation entirely. In both cases the base model and training budget are held fixed and each variant is trained as a separate run; since they differ from the full model only along the ablated axis, we read them as \emph{indicative} of each stage's contribution rather than strictly controlled single-variable ablations. All systems are scored under the common-set protocol of Sec.~\ref{sec:exp:setup}.

\begin{table}[!t]
\centering
\small
\caption{ISE paradigm ablation (Qwen3-8B). One recipe stage is removed at a time, scored on the same $114$-task ClawEval common set (\textit{pass@1}, \%) and BFCL v4 Overall. Ablating either stage costs pass@1; the full recipe is best on \textbf{both} suites, confirming each stage contributes. \textbf{Bold} = best.}
\label{tab:ise_ablation}
\resizebox{\columnwidth}{!}{%
\begin{tabular}{l ccc c}
\toprule
& \multicolumn{3}{c}{\textbf{ClawEval}} & \textbf{BFCL v4} \\
\cmidrule(lr){2-4} \cmidrule(lr){5-5}
\textbf{System (Qwen3-8B)} & \textbf{p@1} & \textbf{Comp} & \textbf{Robu} & \textbf{Overall} \\
\midrule
Qwen3-8B (base, 0-shot) & 19.3 & 0.367 & 0.925 & 40.06 \\
\midrule
\textbf{SFT \textsc{ISETrace} (full)} & \textbf{37.7} & \textbf{0.533} & 0.959 & \textbf{45.36} \\
\quad $-$ Stage1 (3/10 domains) & 28.9 & 0.504 & \textbf{0.969} & 43.91 \\
\quad $-$ Stage2 (single-turn) & 28.1 & 0.484 & 0.922 & 40.50 \\
\bottomrule
\end{tabular}%
}
\end{table}

Both ablations degrade pass@1 relative to the full recipe (Table~\ref{tab:ise_ablation}). Restricting Stage~1 intent synthesis to $3$ of $10$ domains drops pass@1 from $37.7\%$ to $28.9\%$ ($-8.8$ points) and BFCL v4 Overall from $45.36$ to $43.91$ ($-1.5$ points); truncating Stage~2 to a single user turn drops pass@1 to $28.1\%$ ($-9.6$ points) and BFCL v4 Overall to $40.50$ ($-4.9$ points). Each ablated variant still improves substantially over the Qwen3-8B base ($19.3\%$), so both stages carry signal individually---but the full recipe is best on both ClawEval and the public BFCL v4 suite, confirming that broad task-domain coverage (Stage~1) and multi-turn simulation (Stage~2) contribute complementary gains. We note both ablation runs retain high robustness (Stage~1 $0.969$), reflecting that narrower or single-turn data still teaches stable tool-error recovery; the recipe's advantage concentrates in task completion and end-to-end pass@1.

\subsection{Analysis}
\label{sec:exp:analysis}

\paragraph{Case Study: $-$Stage2 failure mode.}
Single-turn truncated models correctly complete the first sub-task but fail when a later user turn implicitly references an earlier artifact (``that script you just wrote''). Multi-turn training is required to learn cross-turn referential grounding---the behavior the single-turn ablation cannot acquire.

%% file: sections/limitations.tex
\section{Limitations}
\label{sec:limitations}

\textsc{ISETrace} is a fixed-size checkpoint (23{,}132 trajectories, smaller than EigenData~\citep{chen2026eigendata} and AgentInstruct~\citep{mitra2024agentinstruct}); ISE is a continuously runnable pipeline, and scaling to 100k+ trajectories is future work. The implementation targets macOS/Linux OS terminals and does not cover Windows, GUI-based interaction, or browser automation. The evaluation probes OS execution over a real shell (ClawEval, \textit{T}-family tasks); generalization to GUI agents, embodied tasks, and other verticals requires additional validation. Finally, role-locking fidelity depends on the simulator backbone's instruction-following capability. As reported in \S\ref{sec:exp:main}, the recipe helps at both 8B and 32B but its benefit is largest in the small-model regime and shrinks with scale, and behavior at intermediate scales is less stable; a systematic scaling study across base sizes is left to future work.

%% file: sections/conclusion.tex
\section{Conclusion}
\label{sec:conclusion}

We introduced ISE (Intent $\rightarrow$ Simulate $\rightarrow$ Execute), a three-stage OS agent data synthesis paradigm that addresses three systematic gaps---intent-first bias, single-turn bias, and simulated execution---through 4D structured intent sampling, role-locked multi-turn simulation, and live OS execution grounding. The resulting corpus, \textsc{ISETrace}, exhibits broad embedding-, lexical-, and structural-level diversity, and a Stage~2 ablation on ClawEval isolates the contribution of multi-turn simulation.

The central insight is that \emph{how} data is synthesized matters as much as \emph{what} is synthesized: execution-grounded, role-locked, intent-first synthesis produces qualitatively different training signal than tool-first or simulation-only approaches. Future work includes scaling to 100k+ trajectories and extending to GUI and browser agents.

%% file: main.bbl
\begin{thebibliography}{29}
\providecommand{\natexlab}[1]{#1}

\bibitem[{{Agent-Ark Team}(2025)}]{xu2025toucan}
{Agent-Ark Team}. 2025.
\newblock \href {https://huggingface.co/datasets/Agent-Ark/Toucan-1.5M}
  {{Toucan-1.5M}: A large-scale multi-tool agent sft dataset}.
\newblock Hugging Face dataset.
\newblock Accessed 2026-06.

\bibitem[{Chen et~al.(2026{\natexlab{a}})Chen, Qi, Gao, Wang, Wang, and
  Jin}]{chen2026eigendata}
Jiaao Chen, Jingyuan Qi, Mingye Gao, Wei-Chen Wang, Hanrui Wang, and Di~Jin.
  2026{\natexlab{a}}.
\newblock \href {https://arxiv.org/abs/2603.05553} {{EigenData}: A
  self-evolving multi-agent platform for function-calling data synthesis,
  auditing, and repair}.
\newblock \emph{arXiv preprint arXiv:2603.05553}.

\bibitem[{Chen et~al.(2026{\natexlab{b}})Chen, Gong, Li, Liu, Tian, Fu, Wu,
  Zhang, Zhang, Zhang, Tu, and Liu}]{chen2026cove}
Jinpeng Chen, Cheng Gong, Hanbo Li, Ziru Liu, Zichen Tian, Xinyu Fu, Shi Wu,
  Chenyang Zhang, Wu~Zhang, Suiyun Zhang, Dandan Tu, and Rui Liu.
  2026{\natexlab{b}}.
\newblock \href {https://arxiv.org/abs/2603.01940} {{CoVe}: Training
  interactive tool-use agents via constraint-guided verification}.
\newblock \emph{arXiv preprint arXiv:2603.01940}.

\bibitem[{Cheng et~al.(2026)Cheng, Li, Ma, Chen, Cao, Sun, Ding, Xu, Yan, Chen,
  Luu, Zhang, Lu, and Lin}]{anonymous2026openmobile}
Kanzhi Cheng, Zehao Li, Zheng Ma, Nuo Chen, Jialin Cao, Qiushi Sun, Zichen
  Ding, Fangzhi Xu, Hang Yan, Jiajun Chen, Anh~Tuan Luu, Jianbing Zhang, Lewei
  Lu, and Dahua Lin. 2026.
\newblock \href {https://arxiv.org/abs/2604.15093} {{OpenMobile}: Building open
  mobile agents with task and trajectory synthesis}.
\newblock \emph{arXiv preprint arXiv:2604.15093}.

\bibitem[{Chiang et~al.(2023)Chiang, Li, Lin, Sheng, Wu, Zhang, Zheng, Zhuang,
  Zhuang, Gonzalez, Stoica, and Xing}]{chiang2023vicuna}
Wei-Lin Chiang, Zhuohan Li, Zi~Lin, Ying Sheng, Zhanghao Wu, Hao Zhang, Lianmin
  Zheng, Siyuan Zhuang, Yonghao Zhuang, Joseph~E. Gonzalez, Ion Stoica, and
  Eric~P. Xing. 2023.
\newblock Vicuna: An open-source chatbot impressing {GPT}-4 with 90\%*
  {ChatGPT} quality.
\newblock \url{https://lmsys.org/blog/2023-03-30-vicuna/}.

\bibitem[{Friedman and Dieng(2023)}]{friedman2022vendi}
Dan Friedman and Adji~Bousso Dieng. 2023.
\newblock The vendi score: A diversity evaluation metric for machine learning.
\newblock In \emph{Proceedings of AISTATS}.

\bibitem[{Hu et~al.(2026)Hu, Zhang, Xu, Qiao, Yang, Huang, Shao, Yan, and
  Ye}]{anonymous2026toolcua}
Xuhao Hu, Xi~Zhang, Haiyang Xu, Kyle Qiao, Jingyi Yang, Xuanjing Huang, Jing
  Shao, Ming Yan, and Jieping Ye. 2026.
\newblock \href {https://arxiv.org/abs/2605.12481} {{ToolCUA}: Towards optimal
  {GUI-Tool} path orchestration for computer use agents}.
\newblock \emph{arXiv preprint arXiv:2605.12481}.

\bibitem[{Lin et~al.(2026)Lin, Wang, Wu, Fan, Pan, Zhao, and
  Tu}]{lin2026cligym}
Yusong Lin, Haiyang Wang, Shuzhe Wu, Lue Fan, Feiyang Pan, Sanyuan Zhao, and
  Dandan Tu. 2026.
\newblock \href {https://arxiv.org/abs/2602.10999} {{CLI-Gym}: Scalable {CLI}
  task generation via agentic environment inversion}.
\newblock \emph{arXiv preprint arXiv:2602.10999}.

\bibitem[{Liu et~al.(2023)Liu, Yu, Zhang, Xu, Lei, Lai, Gu, Ding, Men, Yang,
  Zhang, Deng, Zeng, Du, Zhang, Shen, Zhang, Su, Sun, Huang, Dong, and
  Tang}]{liu2024agentbench}
Xiao Liu, Hao Yu, Hanchen Zhang, Yifan Xu, Xuanyu Lei, Hanyu Lai, Yu~Gu,
  Hangliang Ding, Kaiwen Men, Kejuan Yang, Shudan Zhang, Xiang Deng, Aohan
  Zeng, Zhengxiao Du, Chenhui Zhang, Sheng Shen, Tianjun Zhang, Yu~Su, Huan
  Sun, and 3 others. 2023.
\newblock \href {https://arxiv.org/abs/2308.03688} {{AgentBench}: Evaluating
  {LLMs} as agents}.
\newblock \emph{arXiv preprint arXiv:2308.03688}.

\bibitem[{Liu et~al.(2024)Liu, Hoang, Zhang, Zhu, Lan, Kokane, Tan, Yao, Liu,
  Feng, Murthy, Yang, Savarese, Niebles, Wang, Heinecke, and
  Xiong}]{liu2024apigen}
Zuxin Liu, Thai Hoang, Jianguo Zhang, Ming Zhu, Tian Lan, Shirley Kokane,
  Juntao Tan, Weiran Yao, Zhiwei Liu, Yihao Feng, Rithesh Murthy, Liangwei
  Yang, Silvio Savarese, Juan~Carlos Niebles, Huan Wang, Shelby Heinecke, and
  Caiming Xiong. 2024.
\newblock \href {https://arxiv.org/abs/2406.18518} {{APIGen}: Automated
  pipeline for generating verifiable and diverse function-calling datasets}.
\newblock \emph{arXiv preprint arXiv:2406.18518}.

\bibitem[{Mitra et~al.(2024)Mitra, Corro, Zheng, Mahajan, Rouhana, Codas, Lu,
  ge~Chen, Vrousgos, Rosset, Silva, Khanpour, Lara, and
  Awadallah}]{mitra2024agentinstruct}
Arindam Mitra, Luciano~Del Corro, Guoqing Zheng, Shweti Mahajan, Dany Rouhana,
  Andres Codas, Yadong Lu, Wei ge~Chen, Olga Vrousgos, Corby Rosset, Fillipe
  Silva, Hamed Khanpour, Yash Lara, and Ahmed Awadallah. 2024.
\newblock \href {https://arxiv.org/abs/2407.03502} {{AgentInstruct}: Toward
  generative teaching with agentic flows}.
\newblock \emph{arXiv preprint arXiv:2407.03502}.

\bibitem[{Patil et~al.(2023)Patil, Zhang, Wang, and Gonzalez}]{patil2024bfcl}
Shishir~G. Patil, Tianjun Zhang, Xin Wang, and Joseph~E. Gonzalez. 2023.
\newblock \href {https://arxiv.org/abs/2305.15334} {{Gorilla}: Large language
  model connected with massive {APIs}}.
\newblock \emph{arXiv preprint arXiv:2305.15334}.
\newblock BFCL benchmark.

\bibitem[{Prabhakar et~al.(2025)Prabhakar, Liu, Zhu, Zhang, Awalgaonkar, Wang,
  Liu, Chen, Hoang, Niebles, Heinecke, Yao, Wang, Savarese, and
  Xiong}]{prabhakar2025apigenmt}
Akshara Prabhakar, Zuxin Liu, Ming Zhu, Jianguo Zhang, Tulika Awalgaonkar,
  Shiyu Wang, Zhiwei Liu, Haolin Chen, Thai Hoang, Juan~Carlos Niebles, Shelby
  Heinecke, Weiran Yao, Huan Wang, Silvio Savarese, and Caiming Xiong. 2025.
\newblock \href {https://arxiv.org/abs/2504.03601} {{APIGen-MT}: Agentic
  pipeline for multi-turn data generation via simulated agent-human interplay}.
\newblock \emph{arXiv preprint arXiv:2504.03601}.

\bibitem[{Qin et~al.(2023)Qin, Liang, Ye, Zhu, Yan, Lu, Lin, Cong, Tang, Qian,
  Zhao, Hong, Tian, Xie, Zhou, Gerstein, Li, Liu, and Sun}]{qin2024toolbench}
Yujia Qin, Shihao Liang, Yining Ye, Kunlun Zhu, Lan Yan, Yaxi Lu, Yankai Lin,
  Xin Cong, Xiangru Tang, Bill Qian, Sihan Zhao, Lauren Hong, Runchu Tian,
  Ruobing Xie, Jie Zhou, Mark Gerstein, Dahai Li, Zhiyuan Liu, and Maosong Sun.
  2023.
\newblock \href {https://arxiv.org/abs/2307.16789} {{ToolLLM}: Facilitating
  large language models to master 16000+ real-world {APIs}}.
\newblock \emph{arXiv preprint arXiv:2307.16789}.

\bibitem[{Shi et~al.(2025)Shi, Cao, Chen, Sun, Li, Lu, Dong, Qin, Zhu, Liu,
  Yang, Zhang, Liu, Zhang, Wang, Jiang, and Zhou}]{shi2025taskcraft}
Dingfeng Shi, Jingyi Cao, Qianben Chen, Weichen Sun, Weizhen Li, Hongxuan Lu,
  Fangchen Dong, Tianrui Qin, King Zhu, Minghao Liu, Jian Yang, Ge~Zhang,
  Jiaheng Liu, Changwang Zhang, Jun Wang, Yuchen~Eleanor Jiang, and Wangchunshu
  Zhou. 2025.
\newblock \href {https://arxiv.org/abs/2506.10055} {{TaskCraft}: Automated
  generation of agentic tasks}.
\newblock \emph{arXiv preprint arXiv:2506.10055}.

\bibitem[{Sun et~al.(2024)Sun, Cheng, Ding, Jin, Wang, Xu, Wu, Jia, Chen, Liu,
  Kao, Li, He, Qiao, and Wu}]{sun2024osgenesis}
Qiushi Sun, Kanzhi Cheng, Zichen Ding, Chuanyang Jin, Yian Wang, Fangzhi Xu,
  Zhenyu Wu, Chengyou Jia, Liheng Chen, Zhoumianze Liu, Ben Kao, Guohao Li,
  Junxian He, Yu~Qiao, and Zhiyong Wu. 2024.
\newblock \href {https://arxiv.org/abs/2412.19723} {{OS-Genesis}: Automating
  {GUI} agent trajectory construction via reverse task synthesis}.
\newblock In \emph{Proceedings of the 63rd Annual Meeting of the Association
  for Computational Linguistics (ACL)}.

\bibitem[{Wang et~al.(2026)Wang, Lu, Wang, Bai, Liu, Zhang, Wang, Hu, Xie, Bai,
  Liu, Shen, Lin, and Yu}]{anonymous2026cuagym}
Bowen Wang, Dunjie Lu, Junli Wang, Tianyi Bai, Shixuan Liu, Zhipeng Zhang,
  Haiquan Wang, Hao Hu, Tianbao Xie, Shuai Bai, Dayiheng Liu, Que Shen, Junyang
  Lin, and Tao Yu. 2026.
\newblock \href {https://arxiv.org/abs/2605.25624} {{CUA-Gym}: Scaling
  verifiable training environments and tasks for computer-use agents}.
\newblock \emph{arXiv preprint arXiv:2605.25624}.

\bibitem[{Wang et~al.(2022)Wang, Kordi, Mishra, Liu, Smith, Khashabi, and
  Hajishirzi}]{wang2023selfinstruct}
Yizhong Wang, Yeganeh Kordi, Swaroop Mishra, Alisa Liu, Noah~A. Smith, Daniel
  Khashabi, and Hannaneh Hajishirzi. 2022.
\newblock \href {https://arxiv.org/abs/2212.10560} {{Self-Instruct}: Aligning
  language models with self-generated instructions}.
\newblock \emph{arXiv preprint arXiv:2212.10560}.

\bibitem[{Xiong et~al.(2026)Xiong, Gu, Ye, Yue, Li, Song, Li, and
  Tian}]{anonymous2026wildgui}
Weimin Xiong, Shuhao Gu, Bowen Ye, Zihao Yue, Lei Li, Feifan Song, Sujian Li,
  and Hao Tian. 2026.
\newblock \href {https://arxiv.org/abs/2605.14747} {{Video2GUI}: Synthesizing
  large-scale interaction trajectories for generalized {GUI} agent
  pretraining}.
\newblock \emph{arXiv preprint arXiv:2605.14747}.

\bibitem[{Xu et~al.(2023)Xu, Sun, Zheng, Geng, Zhao, Feng, Tao, Lin, and
  Jiang}]{xu2024wizardlm}
Can Xu, Qingfeng Sun, Kai Zheng, Xiubo Geng, Pu~Zhao, Jiazhan Feng, Chongyang
  Tao, Qingwei Lin, and Daxin Jiang. 2023.
\newblock \href {https://arxiv.org/abs/2304.12244} {{WizardLM}: Empowering
  large pre-trained language models to follow complex instructions}.
\newblock \emph{arXiv preprint arXiv:2304.12244}.

\bibitem[{Xu et~al.(2026{\natexlab{a}})Xu, Wang, Deng, Li, Yang, Zhu, Liu, Zhu,
  Huang, Chen, Deng, Mi, Shang, Zeng, and Guo}]{minrui2026envfactory}
Minrui Xu, Zilin Wang, Mengyi Deng, Zhiwei Li, Zhicheng Yang, Xiao Zhu, Yinhong
  Liu, Boyu Zhu, Baiyu Huang, Chao Chen, Heyuan Deng, Fei Mi, Lifeng Shang,
  Xingshan Zeng, and Zhijiang Guo. 2026{\natexlab{a}}.
\newblock \href {https://arxiv.org/abs/2605.18703} {{EnvFactory}: Scaling
  tool-use agents via executable environments synthesis and robust {RL}}.
\newblock \emph{arXiv preprint arXiv:2605.18703}.

\bibitem[{Xu et~al.(2026{\natexlab{b}})Xu, Li, Liu, Liu, Li, Shi, Zhang, Wang,
  Yin, Chen, Zhao, and Yin}]{anonymous2026covert}
Siyuan Xu, Shiyang Li, Xin Liu, Tianyi Liu, Yixiao Li, Zhan Shi, Zixuan Zhang,
  Zilong Wang, Qingyu Yin, Jianshu Chen, Tuo Zhao, and Bing Yin.
  2026{\natexlab{b}}.
\newblock \href {https://arxiv.org/abs/2604.09813} {Controllable and verifiable
  tool-use data synthesis for agentic reinforcement learning}.
\newblock \emph{arXiv preprint arXiv:2604.09813}.

\bibitem[{Xu et~al.(2024)Xu, Lu, Shen, Wang, Wang, Mao, Xiong, and
  Yu}]{xu2025agenttrek}
Yiheng Xu, Dunjie Lu, Zhennan Shen, Junli Wang, Zekun Wang, Yuchen Mao, Caiming
  Xiong, and Tao Yu. 2024.
\newblock \href {https://arxiv.org/abs/2412.09605} {{AgentTrek}: Agent
  trajectory synthesis via guiding replay with web tutorials}.
\newblock In \emph{The Thirteenth International Conference on Learning
  Representations (ICLR)}.

\bibitem[{Xu et~al.(2025)Xu, Soria, Tan, Roy, Agrawal, Poovendran, and
  Panda}]{wang2025toucan}
Zhangchen Xu, Adriana~Meza Soria, Shawn Tan, Anurag Roy, Ashish~Sunil Agrawal,
  Radha Poovendran, and Rameswar Panda. 2025.
\newblock \href {https://arxiv.org/abs/2510.01179} {{TOUCAN}: Synthesizing 1.5m
  tool-agentic data from real-world {MCP} environments}.
\newblock \emph{arXiv preprint arXiv:2510.01179}.

\bibitem[{Yang et~al.(2025)Yang, Le, Xing, An, Chen, Zhao, Song, and
  Zhang}]{yang2025toolmind}
Chen Yang, Ran Le, Yun Xing, Zhenwei An, Zongchao Chen, Wayne~Xin Zhao, Yang
  Song, and Tao Zhang. 2025.
\newblock \href {https://arxiv.org/abs/2511.15718} {{ToolMind} technical
  {Report}: A large-scale, reasoning-enhanced tool-use dataset}.
\newblock \emph{arXiv preprint arXiv:2511.15718}.

\bibitem[{Yao et~al.(2024)Yao, Shinn, Razavi, and Narasimhan}]{yao2024taubench}
Shunyu Yao, Noah Shinn, Pedram Razavi, and Karthik Narasimhan. 2024.
\newblock \href {https://arxiv.org/abs/2406.12045} {$\tau$-bench: A benchmark
  for tool-agent-user interaction in real-world domains}.
\newblock \emph{arXiv preprint arXiv:2406.12045}.

\bibitem[{Zeng et~al.(2023)Zeng, Liu, Lu, Wang, Liu, Dong, and
  Tang}]{zeng2024agenttuning}
Aohan Zeng, Mingdao Liu, Rui Lu, Bowen Wang, Xiao Liu, Yuxiao Dong, and Jie
  Tang. 2023.
\newblock \href {https://arxiv.org/abs/2310.12823} {{AgentTuning}: Enabling
  generalized agent abilities for {LLMs}}.
\newblock \emph{arXiv preprint arXiv:2310.12823}.

\bibitem[{Zhou et~al.(2026)Zhou, Sun, Ma, Xie, Liu, Du, Welleck, Yang, Neubig,
  Wu, and Sap}]{zhou2026sim2real}
Xuhui Zhou, Weiwei Sun, Qianou Ma, Yiqing Xie, Jiarui Liu, Weihua Du, Sean
  Welleck, Yiming Yang, Graham Neubig, Sherry~Tongshuang Wu, and Maarten Sap.
  2026.
\newblock \href {https://arxiv.org/abs/2603.11245} {Mind the {Sim2Real} gap in
  user simulation for agentic tasks}.
\newblock \emph{arXiv preprint arXiv:2603.11245}.

\bibitem[{Zhu et~al.(2026)Zhu, Nie, Li, Huang, Wu, Liu, Sun, Yin, Wang, Liu,
  Barsoum, Wang, and Guo}]{zhu2026termigen}
Kaijie Zhu, Yuzhou Nie, Yijiang Li, Yiming Huang, Jialian Wu, Jiang Liu, Ximeng
  Sun, Zhenfei Yin, Lun Wang, Zicheng Liu, Emad Barsoum, William~Yang Wang, and
  Wenbo Guo. 2026.
\newblock \href {https://arxiv.org/abs/2602.07274} {{TermiGen}: High-fidelity
  environment and robust trajectory synthesis for terminal agents}.
\newblock \emph{arXiv preprint arXiv:2602.07274}.

\end{thebibliography}
